\documentclass[final]{anthology-ch}         

\usepackage{framed}
\usepackage{xcolor}
\definecolor{burgundy}{RGB}{128,0,32}
\usepackage{booktabs}
\usepackage{float}
\usepackage{graphicx}
\usepackage{multirow}
\usepackage{makecell}
\usepackage[numbers,sort&compress]{natbib}
\usepackage[ruled,vlined]{algorithm2e}
\usepackage{array}
\usepackage[font=scriptsize]{caption}
\usepackage{subcaption} 
\usepackage{tikz}       
\usepackage{siunitx}    

\newcommand{\panel}[3][0.48\linewidth]{%
  \begin{tikzpicture}
    \node[inner sep=0] (img) {\includegraphics[width=#1]{#2}};
    \node[anchor=north west, xshift=-15pt, yshift=-6pt, fill=white, rounded corners=2pt, inner sep=2pt, text=black, font=\bfseries] 
         at (img.north west) {#3};
  \end{tikzpicture}%
}

\title{VBFDD-Agent for Electric Vehicle Battery Fault Detection and Diagnosis: Descriptive Text Modeling of Battery Digital Signals}

\author[1]{Joey Chan}[
  orcid=0009-0004-8100-1546
]

\author[1]{Zhen Chen}[
  orcid=0000-0003-2590-0307
]

\author[1]{Ershun Pan}[
  orcid=0000-0001-6026-9755
]

\affiliation{1}{
  Department of Industrial Engineering and Management, School of Mechanical Engineering, Shanghai Jiao Tong University, Shanghai 200240, China
}


\pubyear{2025}
\pubvolume{1}
\pagestart{1}
\pageend{1}
\conferencename{Proceedings of Conference XXX}
\conferenceeditors{Editor1 Editor2}
\doi{00000/00000}  

\begin{document}

\maketitle

\begin{abstract}
With the rapid proliferation of electric vehicles, the safety and reliability of lithium-ion batteries have become critical concerns. Effective anomaly detection is essential for ensuring safe battery operation. However, as battery systems and operating scenarios become increasingly complex, battery fault diagnosis and maintenance require stronger cross-domain adaptability and human-AI collaboration. Traditional fault detection and diagnosis methods are usually designed for specific scenarios and predefined workflows, making them less effective in complex real-world applications.
To address the scarcity of open-source battery fault report corpora and the lack of unified maintenance knowledge representation, this study proposes a descriptive text modeling approach for battery signal reports. Monitoring signals, statistical features, anomaly records, and state assessment results are transformed into structured and readable natural language descriptions, forming a language corpus for battery health diagnosis and maintenance.
Based on this corpus, we propose VBFDD-Agent, a vehicle battery fault detection and diagnosis agent for automotive-grade battery systems. VBFDD-Agent integrates descriptive battery-state texts, historical case retrieval, local maintenance manuals, and large language model reasoning to generate structured diagnostic results and maintenance recommendations. Experiments show that the proposed framework can accurately perform anomaly monitoring based on descriptive textual representations and provide flexible, efficient, and actionable maintenance suggestions. Expert evaluation further confirms the practical value of the generated recommendations. Overall, VBFDD-Agent extends traditional battery diagnosis from label prediction to interpretable and maintenance-oriented decision support.
\end{abstract}

    \section{Introduction}

In recent years, driven by climate change concerns and environmental challenges, new energy vehicles, especially electric vehicles, have experienced rapid development. 
Among various battery technologies, lithium-ion batteries have been widely used as the core power source of electric vehicles owing to their high energy density and long cycle life\cite{xiong2020research}. 
However, lithium-ion batteries may suffer from various heterogeneous faults, such as anode degradation, cathode degradation, overcharging, overdischarging, and external short circuits\cite{chan2024variational,hesse2017lithium}. 
To anticipate abnormal conditions before they evolve into severe failures, researchers have developed a variety of fault detection and diagnosis (FDD) methods\cite{zhao2026enhancing,xu2025recent}. 
Accurate and robust battery FDD techniques are therefore essential for the development of reliable battery management systems (BMSs)\cite{khaneghah2023fault}.

Traditional battery FDD studies mainly focus on signal processing, feature extraction, state identification, and fault classification\cite{hu2020advanced}. 
These studies aim to identify abnormal states, fault types, or fault probabilities by using voltage, current, temperature, internal resistance, and their derived features\cite{pu2023compound,wang2021lithium}. 
Considerable progress has been made in early fault warning and single-point fault identification. 
Existing methods can generally be classified into knowledge-based, model-based, statistical, machine learning-based, and hybrid approaches\cite{zou2023review}. 
Knowledge- and model-based methods rely on prior expert knowledge, battery physical models, equivalent circuit models, or state observers, and detect faults by analyzing the residuals between estimated and observed values\cite{dey2017model,feng2016online}. 
These methods provide a certain degree of mechanistic interpretability, but they often face high modeling costs and difficulties in parameter updating under complex operating conditions and large-scale applications. 
Statistical methods directly extract fault-sensitive features from operational signals such as voltage, current, and temperature, enabling abnormal state characterization without requiring an accurate system model\cite{xu2019early,wang2017voltage}. 
However, their performance is usually sensitive to noise, load fluctuations, and varying operating conditions. 
Meanwhile, data-driven methods, especially machine learning and deep learning approaches, learn fault discrimination patterns from historical operational data and have enhanced the detection capability in complex nonlinear scenarios, becoming an important direction in current battery FDD research\cite{zhao2017fault,how2019state}.

Despite these advances, existing methods still largely follow a “perception–classification” paradigm\cite{cunneen2020autonomous,de2023eeg}. 
In other words, most models focus on extracting effective features from monitoring data and producing quantitative discrimination results, while paying limited attention to generating comprehensive conclusions for practical operation and maintenance scenarios, such as fault mechanism interpretation, causal-chain analysis, risk-level assessment, and executable maintenance recommendations.
However, for next-generation BMSs and battery operation and maintenance systems, the research objective should not be limited to producing isolated outputs such as state of charge (SOC), state of health (SOH), or fault categories. 
Instead, it should move toward a higher level of intelligence, where the system can understand battery states, identify risks, generate recommendations, and assist in maintenance decision-making \cite{salazar2022industrial,hu2021research}. 
In this context, existing studies still face the following limitations:

\begin{enumerate}

\item Due to the confidentiality requirements of new energy vehicle manufacturers regarding after-sales work orders, fault handling procedures, and maintenance records, textual data containing real-world battery state deviations, fault phenomena, diagnostic processes, and maintenance recommendations are usually difficult to obtain publicly\cite{yuan2015development}. 
Most existing battery FDD studies rely on structured monitoring data, such as voltage, current, temperature, and SOC, to perform anomaly detection or fault classification.
However, limited attention has been paid to how these numerical signals can be further transformed into textual fault descriptions that align with the understanding habits of operation and maintenance personnel. 
This issue is particularly prominent in vehicle-wise traction battery scenarios, where fault manifestations are closely related to vehicle operating conditions, BMS alarm logic, after-sales inspection procedures, and maintenance regulations. 
Therefore, diagnostic results represented only by labels or scores are often insufficient to support explanation, communication, and decision-making in real-world maintenance practices.

\item Existing methods also lack sufficient generality and interactivity. 
Most of them focus on anomaly identification or fault classification itself, and their outputs are usually represented as labels, scores, or alarms, which are difficult to directly transform into readable, interpretable, and actionable maintenance knowledge. 
Moreover, these end-to-end schemes are often designed for specific scenarios and predefined workflows. Their outputs still require further manual analysis before being used in practical maintenance decisions, which hinders seamless interaction with users and dynamic environments. 
With the increasing demand for dynamic cross-domain generalization and human--artificial intelligence (AI) collaborative intelligence in industrial applications \cite{schleiger2024collaborative}, the inefficiency and limited applicability of traditional paradigms have become increasingly evident\cite{xu2025mragent} .
For battery systems, this limitation is particularly critical because battery fault evolution is usually characterized by multi-factor coupling, cross-level propagation, and strong scenario dependence\cite{nazim2026extending}. 
Merely providing fault categories or fault probabilities is often insufficient to directly support operation and maintenance decisions.

\end{enumerate}

\begin{figure*}[!t]
    \centering
    \includegraphics[width=\textwidth]{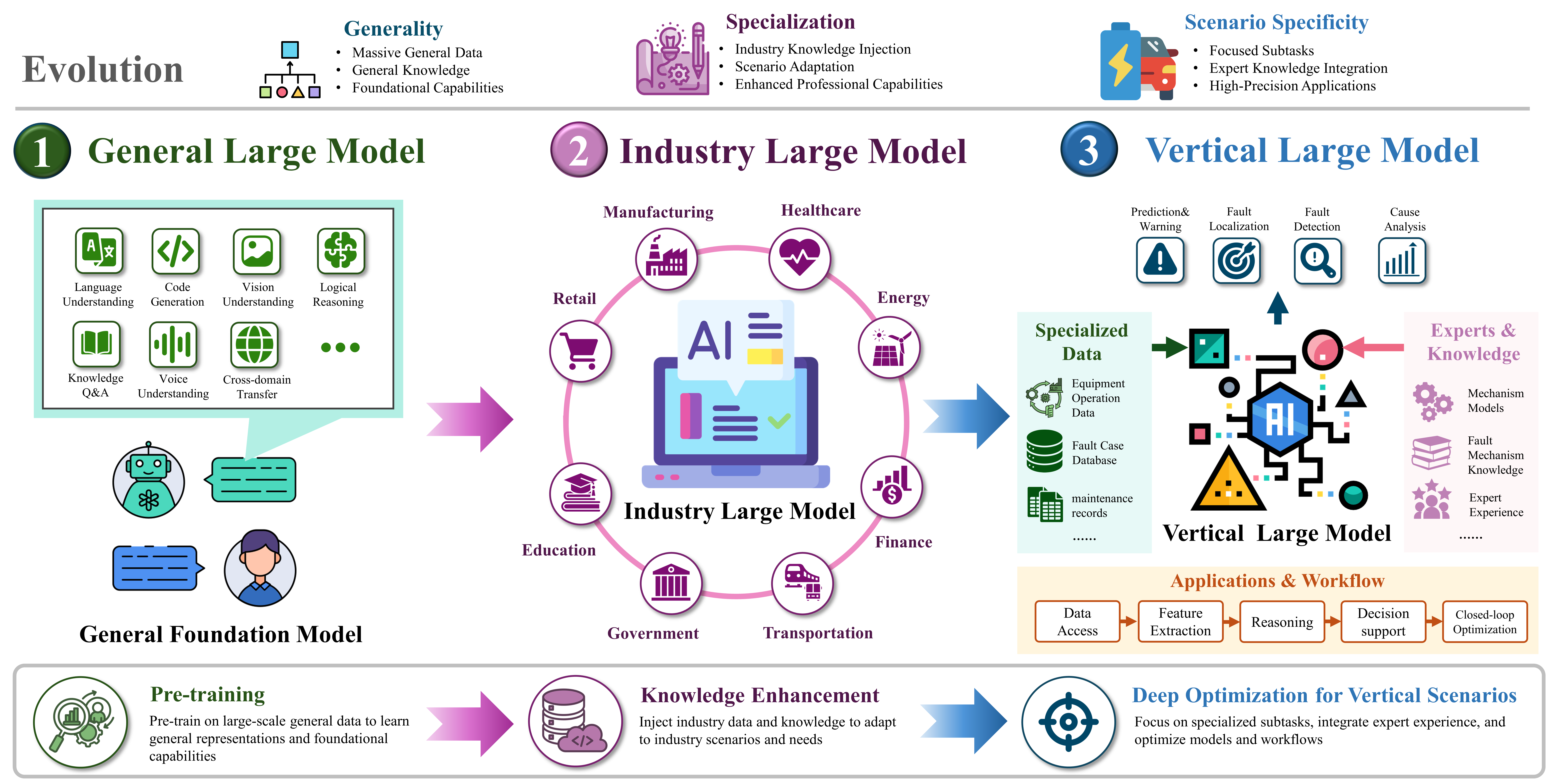}
    \caption{Evolution path from general large models to industry large models and vertical large models for industrial fault diagnosis applications.}
    \label{fig:large_model_evolution}
\end{figure*}

Therefore, how to uniformly represent battery monitoring signals, abnormal states, and maintenance knowledge, and further support intelligent maintenance decision-making, remains an important problem in current battery FDD research. 
In recent years, with the development of the Industrial Agent concept, industrial systems have been gradually evolving from automation tools for local functional optimization toward autonomous decision-making units for full-process collaboration \cite{yu5495908industrial,mavrik2007industrial}. 
Unlike traditional models or tools that only perform a single identification, prediction, or control task, Industrial Agents emphasize the understanding of high-level objectives, autonomous generation of task workflows, and unified orchestration of multiple services and modules\cite{castelfranchi1998modelling}. 
This concept provides a new research direction for battery health operation and maintenance.
Benefiting from the semantic understanding, reasoning traceability, and information retrieval capabilities of large language models (LLMs)\cite{zhao2024explainability}, recent studies have begun to introduce LLMs into equipment health management and fault diagnosis\cite{zhou2026knowledge}. 
For example, Ma et al. \cite{ma2025knowledge} proposed a knowledge graph-enhanced fault diagnosis reasoning framework for analyzing mechanical equipment faults and providing more reliable diagnostic guidance. 
Lin et al. \cite{lin2025fd} proposed a multimodal large model-based method that maps multidimensional time series into a semantic embedding space through modality alignment and then performs inference with an LLM to improve fault classification accuracy. 
Wang et al. \cite{wang2025diagllm} transformed the fault discrimination problem into a visual question answering task and combined envelope spectrum images with domain knowledge to enhance model generalization and interpretability. Although battery health management has made progress in state estimation, degradation prediction, and anomaly warning, its functional modules remain relatively fragmented. 
A unified closed loop from state perception and risk understanding to maintenance handling and scheduling coordination has not yet been fully established \cite{ren}.
Fig.~\ref{fig:large_model_evolution} illustrates this evolution from general large models to industry large models and further to vertical large models, providing the conceptual basis for the proposed VBFDD-Agent.

In this paper, we propose VBFDD-Agent, namely Vehicle Battery Fault Detection and Diagnosis Agent, for vehicle-wise battery fault detection and diagnosis. 
Based on an open-source vehicle battery fault detection dataset, we develop a mechanism-informed descriptive text modeling method for automotive battery state representation. 
The constructed textual corpus bridges the gap between LLMs and FDD tasks. 
Based on battery fault handling manuals from multiple new energy vehicle manufacturers, we further organize battery fault operation and maintenance recommendations and integrate them into the proposed retrieval-augmented Agent framework. 
By considering historical cases and maintenance recommendations, the proposed framework enables the LLM not only to perform complex multi-label monitoring based on descriptive texts, but also to generate scientifically reasonable and practically feasible maintenance recommendations. 
The main contributions of this paper are summarized as follows:

\begin{enumerate}

\item A mechanism-informed signal-to-descriptive-text modeling method is proposed to address the lack of open-source battery operation and maintenance corpora. 
The proposed method transforms tabular numerical signals, which are traditionally processed by neural networks, into descriptive texts that can be more effectively understood by LLMs. 
By incorporating rule-informed preliminary judgments into the textual modeling process, traditional expert rules are effectively embedded into data representation, enabling numerical signal modeling and semantic reasoning to be jointly connected.

\item A retrieval-augmented VBFDD-Agent framework is developed for vehicle battery fault diagnosis and maintenance decision support. 
The proposed Agent integrates historical degradation trajectories, maintenance procedures, case knowledge bases, and real-time operating contexts to support cause tracing, risk grading, impact assessment, and maintenance recommendation generation, rather than merely determining whether a battery is faulty or which fault category it belongs to. 
This framework is expected to help overcome the bottlenecks of isolated information, heavy manual dependence, and long response chains in traditional maintenance modes, thereby promoting battery health management from single-point intelligence to system-level intelligence and from assisted analysis to autonomous decision support.

\item Experimental results on FDD tasks involving ten electric vehicles show that the proposed LLM-based battery FDD framework achieves high diagnostic accuracy, validating the effectiveness of the proposed data modeling method and Agent framework. 
Furthermore, expert evaluation results indicate that the maintenance recommendations generated by VBFDD-Agent are highly recognized by multiple domain experts. 
The constructed corpus will be released to support future research on LLM-based battery fault diagnosis and maintenance.

\end{enumerate}

This article is organized as follows. 
Section 2 presents the problem definition and dataset description. 
Section 3 introduces the proposed method in detail. 
Section 4 verifies the effectiveness of the proposed method. 
Section 5 discusses expert evaluation and model deployment performance. 
Section 6 concludes this article.
    \section{Data Overview}

\begin{figure*}[htbp]
  \centering
  \panel[\linewidth]{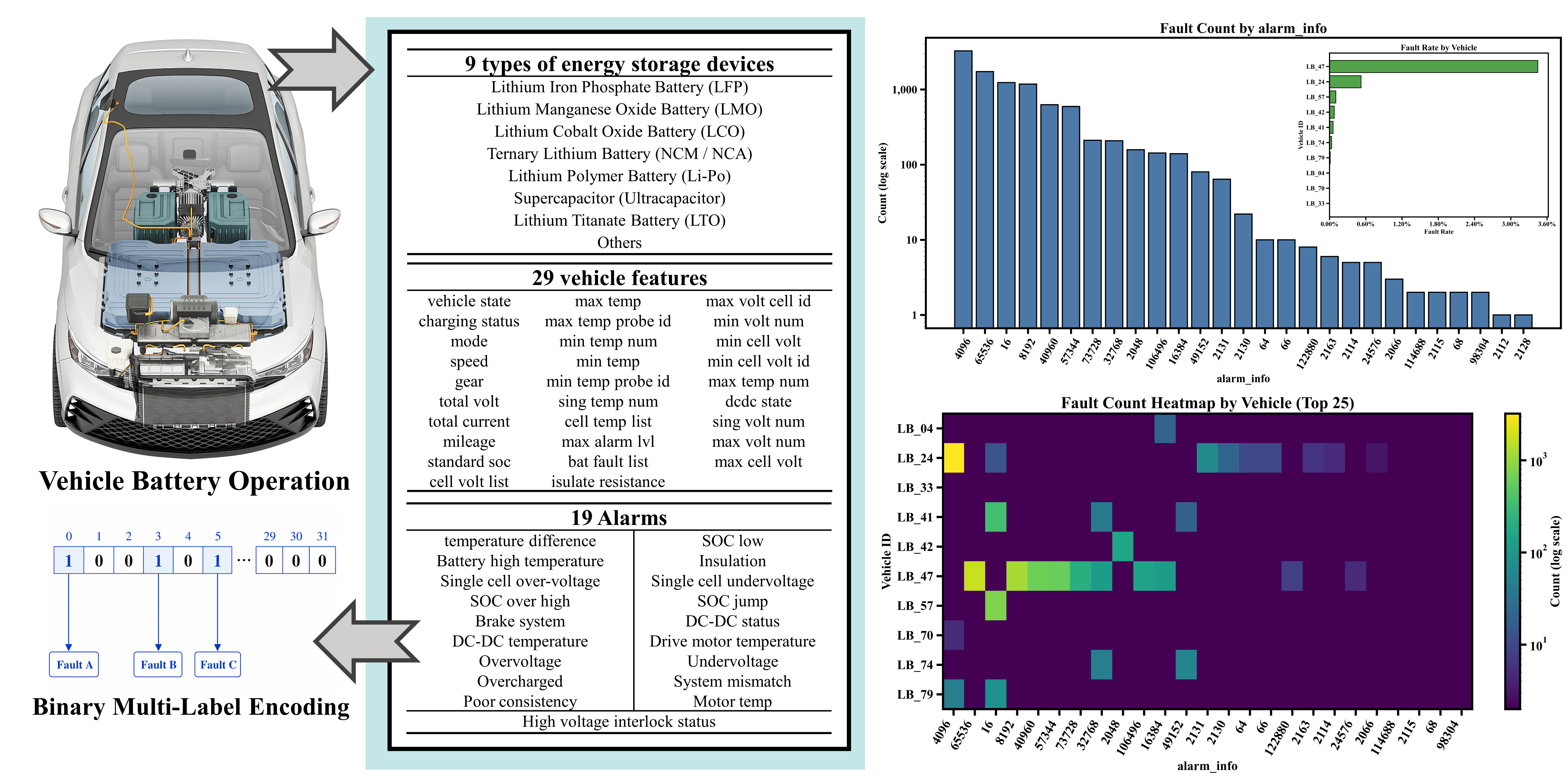}{(a)}\hfill
  \panel[\linewidth]{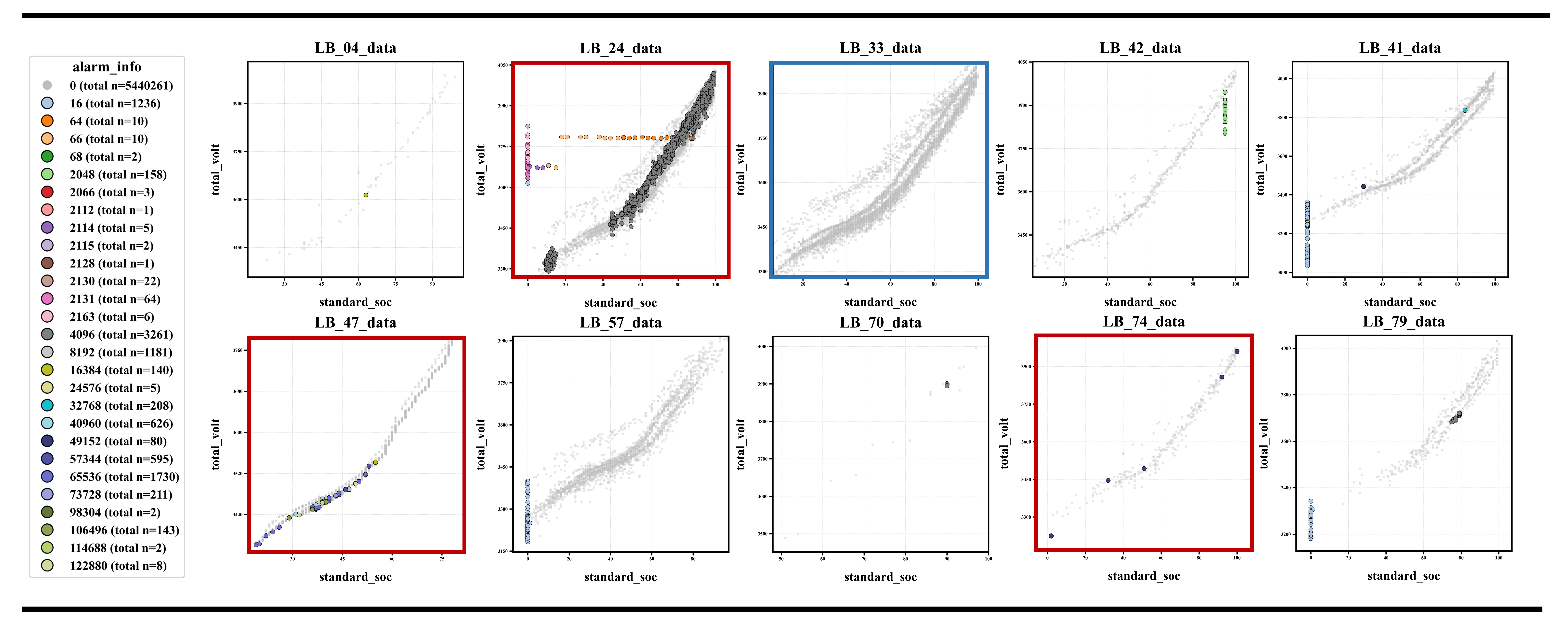}{(b)}
 \caption{Overview of the NDANEV dataset. 
    (a) Data characteristics and fault statistics, showing the sparsity and complexity of fault categories. 
    The binary multi-label encoding allows concurrent fault occurrences within a single record. 
    (b) Sampled record-level visualization based on standard SOC and total voltage. 
    The red boxes indicate vehicles with frequent fault occurrences, while the blue box highlights vehicle LB\_33, for which no fault records are observed.}
    \label{fig:ndanev_soc}
\end{figure*}

The proposed signal-to-text modeling strategy and the VBFDD-Agent framework are built upon raw numerical monitoring signals collected from real-world vehicle operation. 
To this end, we used operational data from ten electric vehicles, denoted as LB\_04, LB\_24, LB\_33, LB\_41, LB\_42, LB\_47, LB\_57, LB\_70, LB\_74, and LB\_79. 
These data were obtained from the National Data Alliance of New Energy Vehicles (NDANEV) \href{http://www.ndanev.com}{http://www.ndanev.com}. 
The datasets cover diverse vehicle-wise operating states, including charging status, driving mode, vehicle speed, start--stop status, and gear position, as well as battery-level monitoring variables, such as voltage, current, SOC, and temperature. 
The involved onboard energy storage system types are also diverse, as shown in Fig.~\ref{fig:ndanev_soc}(a).

The original dataset contains real-world operational records collected over approximately six months from the ten vehicles, with a total of 5,504,260 records. 
After data preprocessing, the dataset contains approximately 3.87 million driving-state records, 1.00 million charging-state records, and 0.65 million stationary-state records. 
A total of 29 monitoring variables are selected as the raw input features for subsequent descriptive text modeling. 
These variables cover a wide range of operating conditions, including SOC values from 0\% to 100\%, vehicle speeds from 0 km/h to 220 km/h, different charging and discharging states, and multiple gear positions. 
The recorded temperature-related values range from -40$^\circ$C to 210$^\circ$C, reflecting both normal operating conditions and extreme abnormal measurements. 
In addition, the dataset contains at least 19 types of abnormal conditions, such as high battery temperature alarms and brake system alarms.

The vehicle faults are encoded as binary multi-label vectors, where one data record may correspond to multiple simultaneous fault states. 
For visualization, the binary fault codes are converted into decimal values and displayed in Fig.~\ref{fig:ndanev_soc}(a). 
According to the fault-count statistics, LB\_24 and LB\_47 contain the largest numbers of abnormal samples, while the heatmap indicates that LB\_47 covers the most diverse fault categories. 
From an overall perspective, abnormal samples are extremely scarce in the dataset. 
Only 10,024 abnormal records are observed, accounting for approximately 0.18\% of the entire dataset. 
Such severe class imbalance poses substantial challenges for conventional FDD models in learning reliable fault-discriminative patterns.

Furthermore, the scatter visualization based on total voltage and normalized SOC in Fig.~\ref{fig:ndanev_soc}(b) shows that most fault samples cannot be clearly separated using simple handcrafted features. 
This observation further suggests that vehicle battery faults are often coupled with complex operating conditions and cannot be effectively characterized by isolated numerical indicators alone. 
Therefore, it is necessary to construct structured descriptive texts by integrating monitoring signals, statistical features, abnormal labels, and state assessment results, so as to provide a more suitable representation for LLM-based fault reasoning and maintenance decision support.

    

    \section{Method}
\label{sec:method}

In this section, we describe the proposed VBFDD-Agent framework in detail. The overall methodology consists of two key components: Mechanism-informed Descriptive Text Modeling and LLM-empowered Electric Vehicle Battery Fault Detection and Diagnosis Agent. 
Specifically, the first component constructs descriptive textual representations from raw vehicle battery signals by incorporating mechanism-informed rules and statistical state information. 
The second component further utilizes large language models, historical cases, and maintenance knowledge to support multi-label fault diagnosis and actionable maintenance decision-making.

\subsection{Mechanism-informed Descriptive Text Modeling}

To bridge raw vehicle battery signals and LLM-based fault reasoning, this study develops a mechanism-informed descriptive text modeling strategy. 
Instead of directly feeding tabular numerical records into the LLM, the proposed strategy transforms each vehicle operation record into a structured natural language description. 
The generated text summarizes the vehicle operating state, battery electrical behavior, thermal distribution, insulation condition, and mechanism-oriented risk interpretation. 
In this way, the numerical monitoring signals are converted into semantically interpretable descriptions that are more suitable for LLM understanding and reasoning.
The descriptive text is generated only from physical monitoring variables, including vehicle speed, total voltage, total current, mileage, SOC, maximum and minimum cell voltages, maximum and minimum temperatures, and insulation resistance. 
The proposed text modeling process follows a leakage-free design. 
Fault-indicative fields, including alarm information, maximum alarm level, and battery fault lists, are excluded from the description generation process and are only retained as output labels for subsequent FDD evaluation.

\begin{figure*}[!t]
\centering

\includegraphics[width=\textwidth]{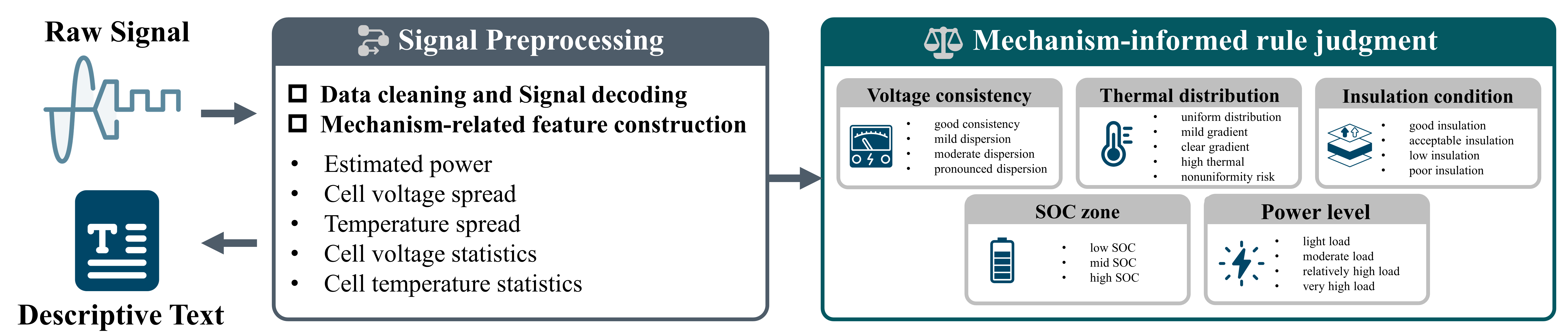}

\vspace{1.5mm}
\centering

\tiny
\renewcommand{\arraystretch}{1.25}
\begin{tabular}{
>{\raggedright\arraybackslash}m{0.10\textwidth}
>{\raggedright\arraybackslash}m{0.15\textwidth}
>{\raggedright\arraybackslash}m{0.24\textwidth}
>{\raggedright\arraybackslash}m{0.43\textwidth}
}
\hline
\textbf{Text segment} 
& \textbf{Input variables} 
& \textbf{Mechanism-informed interpretation} 
& \textbf{Text template} \\
\hline

Vehicle operating state 
& Speed, SOC, total voltage, total current, estimated power, mileage 
& Describes the instantaneous operating condition and load level of the vehicle battery system. 
& At this moment, the vehicle shows the following operating characteristics: speed is about [speed] km/h, SOC is about [SOC]\%, total voltage is about [voltage] V, total current is about [current] A, estimated power is about [power] kW, and mileage is about [mileage] km; [power-level description]. \\

Cell-voltage consistency 
& Maximum cell voltage, minimum cell voltage, cell voltage spread 
& Reflects voltage consistency among cells and indicates possible capacity dispersion, resistance divergence, or polarization heterogeneity. 
& The maximum and minimum cell voltages are about [max-cell-voltage] V and [min-cell-voltage] V, respectively, with a spread of about [voltage-spread] mV; [voltage-consistency description]. \\

Thermal distribution 
& Maximum temperature, minimum temperature, temperature spread 
& Characterizes thermal uniformity and indicates possible nonuniform heat generation or cooling pathways. 
& The maximum and minimum temperatures are about [max-temperature] $^\circ$C and [min-temperature] $^\circ$C, respectively, with a spread of about [temperature-spread] $^\circ$C; [thermal-distribution description]. \\

Insulation condition 
& Insulation resistance 
& Reflects the high-voltage insulation state and potential leakage or insulation-aging risk. 
& The insulation resistance is about [insulation-resistance] k$\Omega$, and [insulation-condition description]. \\

SOC-related state 
& SOC 
& Provides the operating range context for interpreting voltage dispersion and potential degradation behavior. 
& [SOC-zone description]. \\

Mechanism-oriented interpretation 
& SOC, voltage spread, temperature spread, insulation resistance, estimated power 
& Integrates scenario-dependent mechanism rules to summarize possible risk signatures. 
& Mechanism-oriented interpretation: [risk notes]. \\

\hline
\end{tabular}
\caption{Illustration and associated mechanism-informed text template generated from vehicle battery signals.\\
\textbf{Sub-Table:} Template of the mechanism-informed descriptive text generated from vehicle battery signals.}
\label{fig:signal_text_flowchart}
\end{figure*}

Before text generation, the raw signals are decoded and converted into physically meaningful variables. 
Vehicle speed, total voltage, mileage, cell voltage, current, and temperature are converted according to their corresponding measurement scales or encoding rules. 
Based on these variables, several mechanism-related features are constructed, including estimated power, cell voltage spread, temperature spread, cell voltage statistics, and cell temperature statistics. 
These features provide the physical basis for subsequent state interpretation.

Mechanism-informed rules are then embedded into the descriptive text modeling process. 
Specifically, cell voltage spread is used to characterize voltage consistency and potential heterogeneity in capacity, internal resistance, or polarization. 
Temperature spread is used to describe thermal uniformity and possible nonuniform heat generation or cooling. 
Insulation resistance is used to reflect high-voltage insulation safety and leakage risk. 
SOC and power levels are further considered to identify scenario-dependent risk signatures, since voltage dispersion under low or high SOC and high-power operation may expose more pronounced degradation or polarization behaviors. 
The detailed rule-based judgment process is illustrated in Fig.~\ref{fig:signal_text_flowchart}.

Finally, the mechanism-informed features and rule-based interpretations are organized into a unified textual template. 
Each generated description consists of five parts: vehicle operating state, cell-voltage consistency, thermal distribution, insulation condition, and mechanism-oriented interpretation. The template used for descriptive text generation is summarized in Sub-Table of Fig. \ref{fig:signal_text_flowchart}. 
Through this process, each numerical battery record is transformed into a structured description that preserves the physical meaning of the original signals while improving the readability and reasoning compatibility for LLM-based FDD.

\subsection{LLM-empowered Electric Vehicle Battery Fault Detection and Diagnosis Agent}

Based on the mechanism-informed descriptive texts constructed in the previous subsection, this study further develops an LLM-empowered electric vehicle battery fault detection and diagnosis Agent, namely VBFDD-Agent. 
The proposed Agent consists of three key modules: historical case retrieval, alarm-code prediction, and retrieval-augmented LLM reasoning. 
Instead of directly asking the LLM to infer fault labels from raw numerical signals, VBFDD-Agent first retrieves similar historical battery states from the constructed corpus, obtains a preliminary multi-label alarm prediction through similarity-weighted voting, and then combines the predicted alarm information with retrieved maintenance knowledge to generate structured diagnostic and disposal outputs.

\begin{figure*}[!t]
    \centering
    \includegraphics[width=0.5\textwidth]{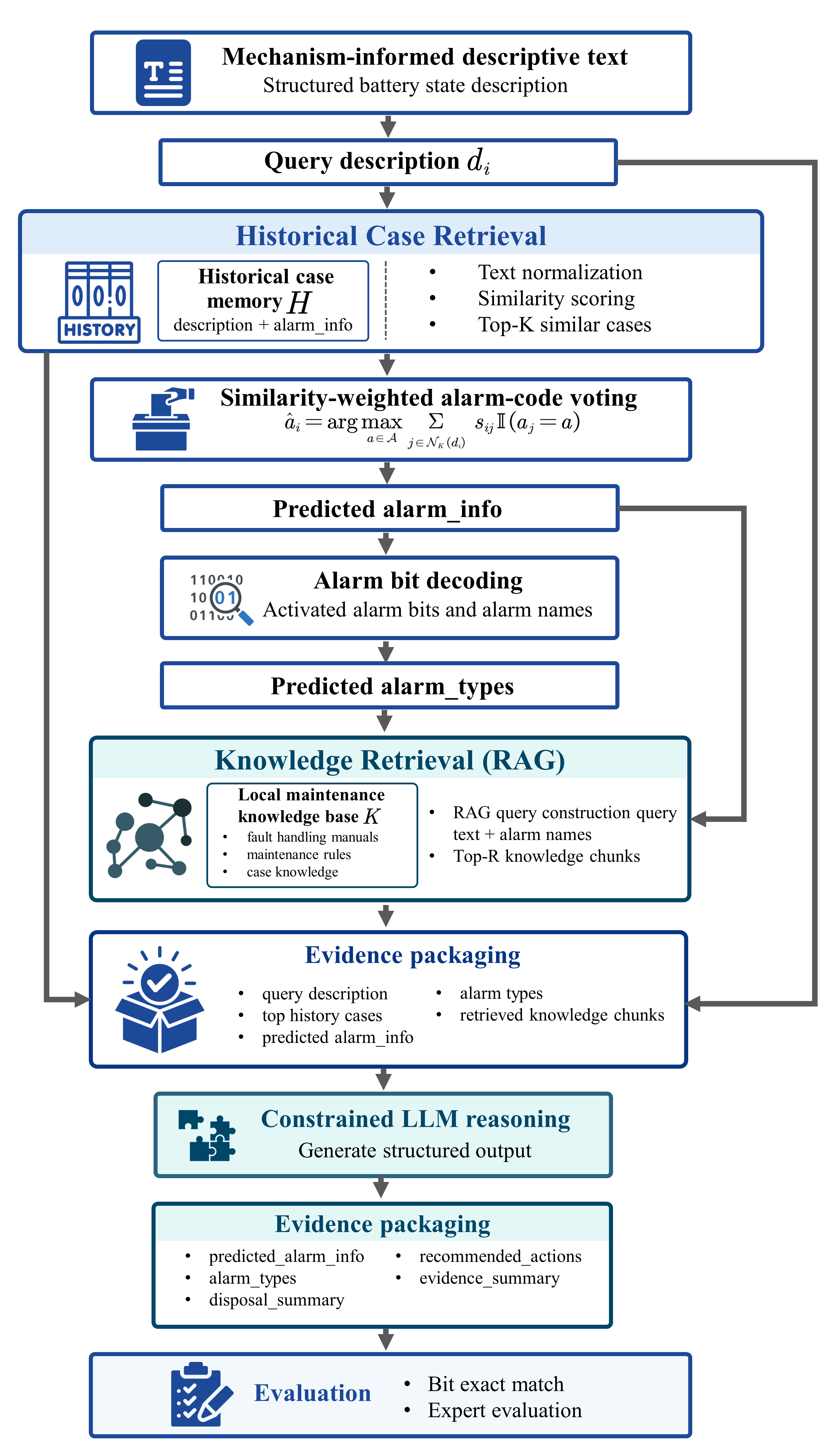}
    \caption{The flowchart of VBFDD-Agent.
        The Agent combines historical case retrieval and maintenance-knowledge retrievalto support multi-label diagnosis and actionable recommendations.}
    \label{fig:agent}
\end{figure*}

Let $\mathbf{x}_i$ denote the raw monitoring record of the $i$-th sample, and let $d_i=\mathcal{T}(\mathbf{x}_i)$ be the corresponding descriptive text generated by the mechanism-informed text modeling function $\mathcal{T}(\cdot)$. 
The alarm state of each sample is represented by an integer alarm code $a_i$, which is further decoded into a binary multi-label vector, as shown in Eq.~\ref{eq:vbfdd_multilabel_vector}.

\begin{equation}
\mathbf{y}_i = [y_{i,0}, y_{i,1}, \ldots, y_{i,B-1}]^\top ,
\label{eq:vbfdd_multilabel_vector}
\end{equation}
where $B$ denotes the number of predefined alarm bits. 
The relationship between the integer alarm code and the multi-label vector is formulated in Eq.~\ref{eq:vbfdd_alarm_code_decoding}.
\begin{equation}
a_i = \sum_{b=0}^{B-1} 2^b y_{i,b}, \quad y_{i,b} \in \{0,1\}.
\label{eq:vbfdd_alarm_code_decoding}
\end{equation}
Equivalently, $y_{i,b}=1$ indicates that the $b$-th alarm type is activated in the current record.

For a query description $d_i$, VBFDD-Agent retrieves similar historical cases from the historical case memory $\mathcal{H}=\{(d_j,a_j)\}_{j=1}^{N_h}$. 
The text similarity score between the query description and each historical description is calculated according to Eq.~\ref{eq:vbfdd_text_similarity}.
\begin{equation}
s_{ij} = \operatorname{Sim}\left(\phi(d_i), \phi(d_j)\right),
\label{eq:vbfdd_text_similarity}
\end{equation}
where $\phi(\cdot)$ denotes text normalization and $\operatorname{Sim}(\cdot)$ denotes the text similarity function. 
Based on the similarity scores, the top-$K$ most similar historical cases are selected using Eq.~\ref{eq:vbfdd_topk_cases}.
\begin{equation}
\mathcal{N}_K(d_i)=\operatorname{TopK}_{j \in \mathcal{H}}(s_{ij}).
\label{eq:vbfdd_topk_cases}
\end{equation}

The preliminary alarm code is predicted by similarity-weighted voting, as defined in Eq.~\ref{eq:vbfdd_weighted_voting}.
\begin{equation}
\hat{a}_i =
\arg\max_{a \in \mathcal{A}}
\sum_{j \in \mathcal{N}_K(d_i)}
s_{ij}\mathbb{I}(a_j=a),
\label{eq:vbfdd_weighted_voting}
\end{equation}
where $\mathcal{A}$ is the set of alarm codes in the historical case memory and $\mathbb{I}(\cdot)$ is the indicator function. 
The predicted integer alarm code $\hat{a}_i$ is decoded into activated alarm bits and corresponding alarm names, forming a structured preliminary diagnosis.

To further generate maintenance recommendations, VBFDD-Agent incorporates a retrieval-augmented generation mechanism. 
Given the query description $d_i$ and the decoded alarm names $\psi(\hat{a}_i)$, relevant maintenance knowledge chunks are retrieved from a local knowledge base $\mathcal{K}$ according to Eq.~\ref{eq:vbfdd_knowledge_retrieval}.
\begin{equation}
\mathcal{R}_R(d_i,\hat{a}_i)
=
\operatorname{TopR}_{c \in \mathcal{K}}
\operatorname{Sim}\left(
\phi(d_i \oplus \psi(\hat{a}_i)), \phi(c)
\right),
\label{eq:vbfdd_knowledge_retrieval}
\end{equation}
where $\oplus$ denotes text concatenation and $\mathcal{R}_R$ represents the top-$R$ retrieved knowledge chunks. These chunks usually contain fault handling procedures, maintenance rules, and operation recommendations.

The final evidence package provided to the LLM is formulated in Eq.~\ref{eq:vbfdd_evidence_package}.
\begin{equation}
\mathcal{E}_i =
\left\{
d_i,\hat{a}_i,\psi(\hat{a}_i),
\mathcal{N}_K(d_i),
\mathcal{R}_R(d_i,\hat{a}_i)
\right\}.
\label{eq:vbfdd_evidence_package}
\end{equation}
With this evidence package, the LLM is constrained to generate a structured JSON output, as shown in Eq.~\ref{eq:vbfdd_llm_output}.
\begin{equation}
\mathbf{o}_i =
\mathcal{G}_{\theta}(\mathcal{E}_i),
\label{eq:vbfdd_llm_output}
\end{equation}
where $\mathcal{G}_{\theta}$ denotes the LLM-based reasoning module. The output $\mathbf{o}_i$ contains the predicted alarm code, activated alarm types, disposal summary, recommended actions, and evidence summary. In this way, the proposed Agent not only performs multi-label battery fault detection, but also transforms the diagnostic result into interpretable and executable maintenance knowledge.

    \section{Experiments}

\subsection{Experiment Setup}

\begin{figure*}[!t]
\centering

\includegraphics[width=0.75\textwidth]{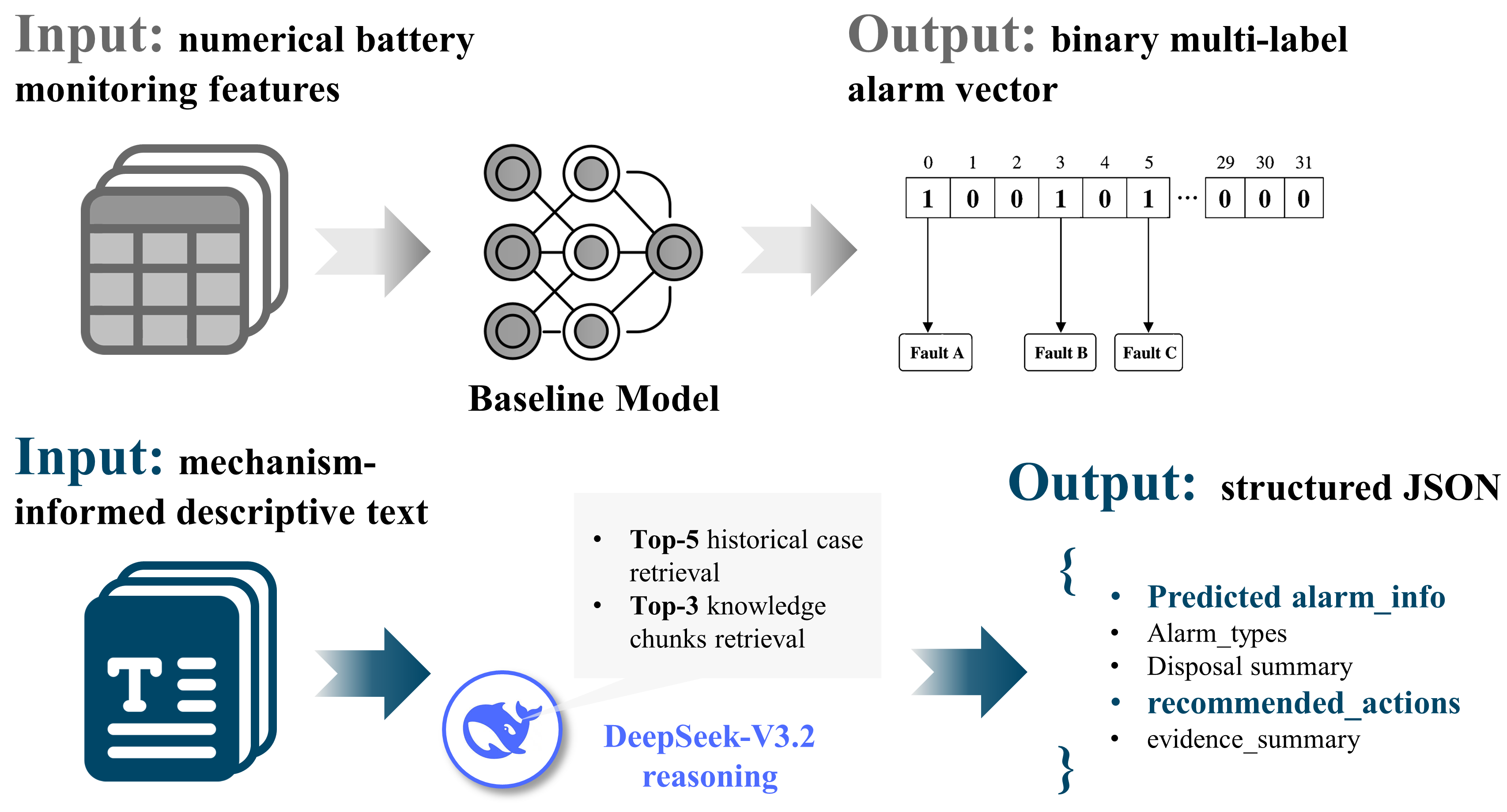}

\vspace{1.5mm}
\centering

\tiny
\renewcommand{\arraystretch}{1.25}

\begin{tabular}{
>{\raggedright\arraybackslash}m{0.13\linewidth}
>{\raggedright\arraybackslash}m{0.68\linewidth}
}
\hline
\textbf{Method} & \textbf{Description} \\
\hline
\textbf{OvR-LR} \cite{alhessi2015swatac} & One-vs-Rest logistic regression for independent alarm-bit classification. \\
\textbf{OvR-SVM} \cite{hong2008probabilistic} & One-vs-Rest linear SVM for multi-label alarm prediction. \\
\textbf{MO-LR} \cite{borchani2015survey} & Multi-output logistic regression for simultaneous alarm-bit prediction. \\
\textbf{CC-LR} \cite{read2011classifier} & Classifier chain with logistic regression to model label dependencies. \\
\textbf{ST-LR} \cite{oymak2020statistical} & Self-training logistic regression with partially masked labels and pseudo-labeling. \\
\textbf{VBFDD-Agent} & Proposed LLM-based Agent using descriptive texts, historical case retrieval, local knowledge retrieval, and structured reasoning. \\
\hline
\multicolumn{2}{>{\raggedright\arraybackslash}p{0.81\linewidth}}{
\textbf{Parameter settings:} Logistic regression: 2000 iterations; linear SVM: 5000 iterations; self-training: 70\% unlabeled labels with a confidence threshold of 0.8.
} \\
\hline
\end{tabular}

\caption{Overall experimental setup of the proposed study.\textbf{Sub-Table:} Compared methods used in the experiments.
}
\label{fig:exp_setup_overview}
\end{figure*}

The overall experimental setup is summarized in Fig.~\ref{fig:exp_setup_overview}. 
For the conventional baselines, numerical battery monitoring features are used as inputs and binary multi-label alarm vectors are used as outputs. 
For the proposed VBFDD-Agent, mechanism-informed descriptive texts are taken as the input, and the Agent further integrates historical case retrieval, local maintenance knowledge retrieval, and DeepSeek-V3.2 reasoning to generate structured diagnostic outputs. 
The lower sub-table in Fig.~\ref{fig:exp_setup_overview} summarizes the compared methods, and key implementation settings in the experiments.



\subsection{Vehicle-wise Anomaly Detection}

\begin{figure*}[htbp]
  \centering
  \panel[0.45\linewidth]{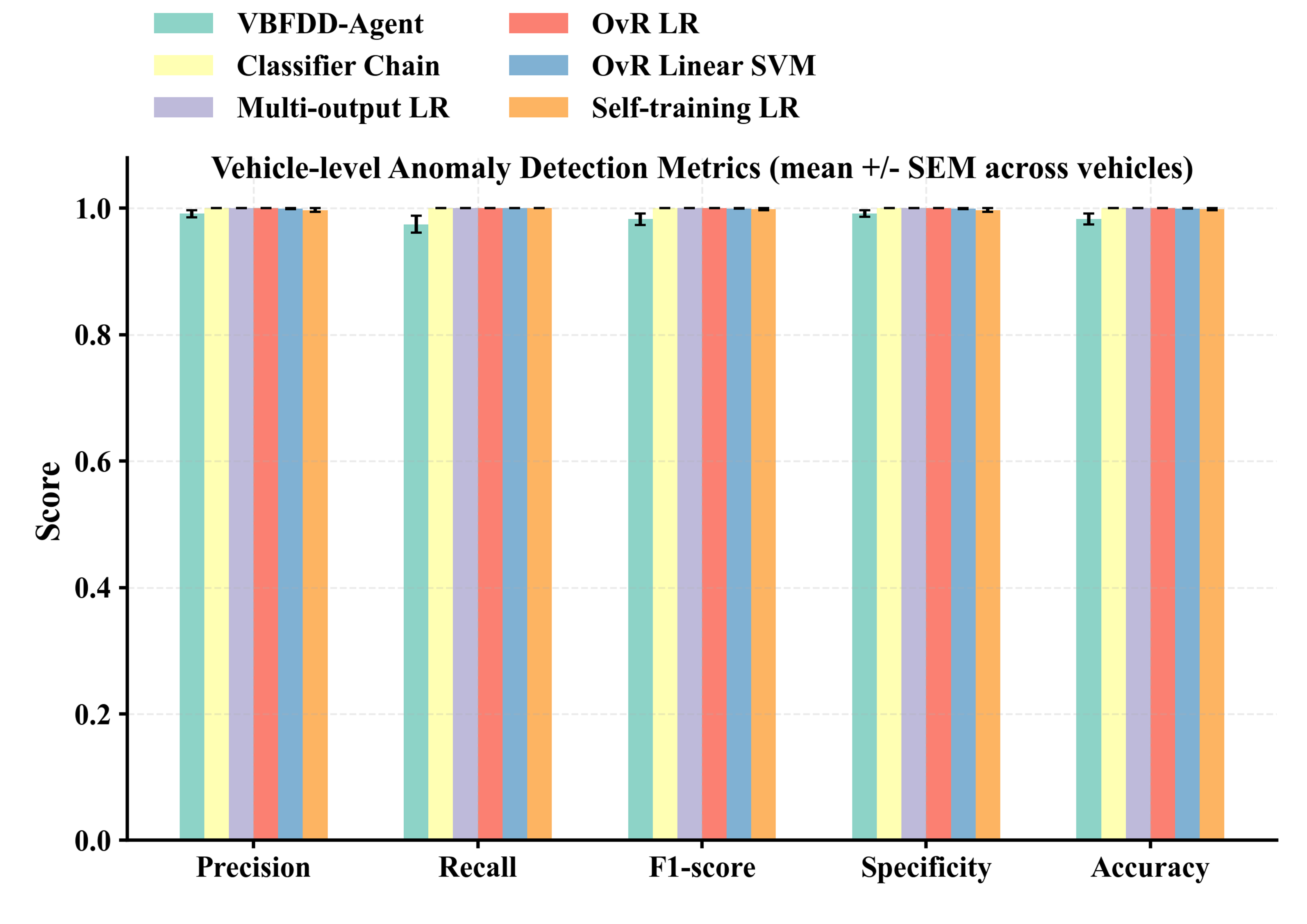}{(a)}\hfill
  \panel[0.45\linewidth]{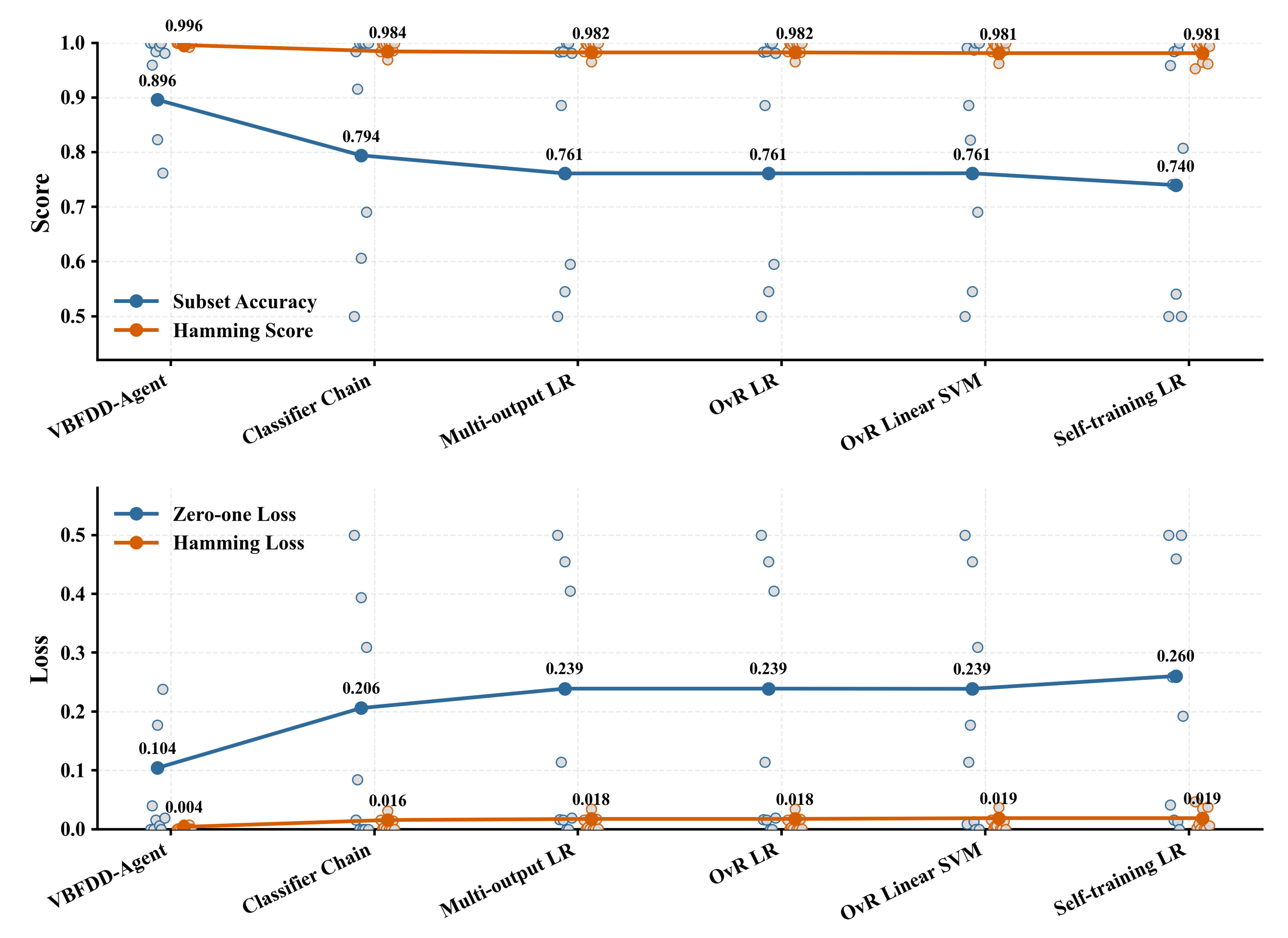}{(b)}
\caption{ 
(a) vehicle-wise anomaly detection results, showing that all compared methods achieve competitive performance in the binary classification task. 
(b) Overall performance of different methods in multi-label fault monitoring, where gray scatter points denote vehicle-wise metric values and orange/blue scatter points denote the mean performance across vehicles.}
    \label{fig:Anomaly_Detection}
\end{figure*}

Before evaluating the fine-grained multi-label fault detection performance, we first conduct a vehicle-wise anomaly detection experiment as a preliminary analysis. 
This task aims to examine whether different methods can distinguish abnormal vehicle battery states from normal operating conditions at a coarse-grained level.
As shown in Fig.~\ref{fig:Anomaly_Detection}(a), all compared methods achieve near-perfect performance in the binary anomaly detection task. 
This indicates that, when the objective is simplified to identifying whether a record is abnormal, both conventional numerical-feature-based models and the proposed VBFDD-Agent can effectively capture the difference between normal and abnormal battery operating states. 
The high performance also suggests that the abnormal samples in the dataset generally exhibit distinguishable signal patterns at the vehicle level.

It should be noted that the proposed VBFDD-Agent follows a training-free diagnostic paradigm, relying on mechanism-informed descriptive texts, historical case retrieval, and LLM-based reasoning rather than task-specific supervised model training. 
Therefore, its binary anomaly detection performance is slightly lower than that of some fully trained baselines in certain vehicle-wise cases. 
However, this result does not indicate a limitation of the proposed framework in practical FDD scenarios. On the contrary, when the task is extended from binary anomaly detection to multi-label fault monitoring, where multiple concurrent faults and more complex alarm patterns need to be identified, 
VBFDD-Agent demonstrates a clear advantage over the compared methods, as shown in Fig.~\ref{fig:Anomaly_Detection}(b).

Overall, the vehicle-wise anomaly detection results show that coarse-grained abnormal state identification is relatively easy for all methods. The more critical challenge lies in accurately identifying complex multi-label fault states and further transforming diagnostic results into interpretable and actionable maintenance recommendations. 
Therefore, the following subsections focus on multi-label fault detection performance and the quality of operation and maintenance suggestions generated by VBFDD-Agent.

\subsection{Multi-label Fault Detection}

After the vehicle-wise binary anomaly detection analysis, we further evaluate the proposed method under a more challenging multi-label fault detection setting. Different from binary anomaly detection, multi-label fault detection requires the model not only to determine whether a vehicle record is abnormal, but also to identify which alarm bits are activated simultaneously. 
This task is more consistent with real-world battery operation and maintenance scenarios, since battery faults often do not occur in isolation. Instead, they may be caused by the coupling of electrical behavior, thermal management, insulation condition, and vehicle-wise operating states, and are usually manifested as concurrent alarm labels.

\begin{table}[htbp]
  \centering
  \caption{Vehicle-wise multi-label fault detection performance.}
  \label{tab:multilabel_prf_jaccard_vehiclewise}
      \resizebox{\textwidth}{!}{
  \begin{tabular}{llccccccccc}
    \toprule
    Metric & Method & LB\_04 & LB\_24 & LB\_41 & LB\_42 & LB\_47 & LB\_57 & LB\_70 & LB\_74 & LB\_79 \\
    \midrule
    \multirow{6}{*}{Micro-Precision} & VBFDD-Agent & \textbf{1.000} & 0.961 & \textbf{1.000} & \textbf{1.000} & \textbf{0.888} & \textbf{1.000} & \textbf{1.000} & \textbf{0.960} & \textbf{1.000} \\
     & \textbf{CC-LR} \cite{read2011classifier}  & 0.500 & 0.958 & 0.861 & \textbf{1.000} & 0.425 & \textbf{1.000} & \textbf{1.000} & 0.646 & \textbf{1.000} \\
     & \textbf{MO-LR} \cite{borchani2015survey} & 0.500 & 0.957 & 0.821 & 0.969 & 0.405 & \textbf{1.000} & \textbf{1.000} & 0.589 & 0.963 \\
     & \textbf{OvR-LR}\cite{alhessi2015swatac}  & 0.500 & 0.957 & 0.821 & 0.969 & 0.405 & \textbf{1.000} & \textbf{1.000} & 0.589 & 0.963 \\
     & \textbf{OvR-SVM} \cite{hong2008probabilistic} & 0.500 & \textbf{0.975} & 0.763 & 0.738 & 0.387 & 0.975 & \textbf{1.000} & 0.647 & \textbf{1.000} \\
     & \textbf{ST-LR} \cite{oymak2020statistical} & 0.250 & 0.912 & 0.649 & 0.969 & 0.400 & 0.975 & \textbf{1.000} & 0.393 & 0.722 \\
    \midrule
    \multirow{6}{*}{Micro-F1} & VBFDD-Agent & \textbf{1.000} & 0.961 & \textbf{1.000} & 0.984 & \textbf{0.838} & 0.994 & \textbf{1.000} & \textbf{0.828} & 0.980 \\
     & \textbf{CC-LR} \cite{read2011classifier}  & 0.667 & 0.974 & 0.926 & \textbf{1.000} & 0.575 & \textbf{1.000} & \textbf{1.000} & 0.765 & \textbf{1.000} \\
     & \textbf{MO-LR} \cite{borchani2015survey} & 0.667 & 0.976 & 0.902 & 0.984 & 0.576 & \textbf{1.000} & \textbf{1.000} & 0.742 & 0.981 \\
     & \textbf{OvR-LR}\cite{alhessi2015swatac}  & 0.667 & 0.976 & 0.902 & 0.984 & 0.576 & \textbf{1.000} & \textbf{1.000} & 0.742 & 0.981 \\
     & \textbf{OvR-SVM} \cite{hong2008probabilistic} & 0.667 & \textbf{0.985} & 0.866 & 0.849 & 0.558 & 0.987 & \textbf{1.000} & 0.786 & \textbf{1.000} \\
     & \textbf{ST-LR} \cite{oymak2020statistical} & 0.400 & 0.952 & 0.787 & 0.984 & 0.572 & 0.987 & \textbf{1.000} & 0.564 & 0.839 \\
    \midrule
    \multirow{6}{*}{Micro-Jaccard} & VBFDD-Agent & \textbf{1.000} & 0.925 & \textbf{1.000} & 0.968 & \textbf{0.721} & 0.987 & \textbf{1.000} & \textbf{0.706} & 0.962 \\
     & \textbf{CC-LR} \cite{read2011classifier}  & 0.500 & 0.950 & 0.861 & \textbf{1.000} & 0.403 & \textbf{1.000} & \textbf{1.000} & 0.620 & \textbf{1.000} \\
     & \textbf{MO-LR} \cite{borchani2015survey} & 0.500 & 0.954 & 0.821 & 0.969 & 0.405 & \textbf{1.000} & \textbf{1.000} & 0.589 & 0.963 \\
     & \textbf{OvR-LR}\cite{alhessi2015swatac}  & 0.500 & 0.954 & 0.821 & 0.969 & 0.405 & \textbf{1.000} & \textbf{1.000} & 0.589 & 0.963 \\
     & \textbf{OvR-SVM} \cite{hong2008probabilistic} & 0.500 & \textbf{0.971} & 0.763 & 0.738 & 0.387 & 0.975 & \textbf{1.000} & 0.647 & \textbf{1.000} \\
     & \textbf{ST-LR} \cite{oymak2020statistical} & 0.250 & 0.908 & 0.649 & 0.969 & 0.400 & 0.975 & \textbf{1.000} & 0.393 & 0.722 \\
    \bottomrule
  \end{tabular}}
\end{table}

Table~\ref{tab:multilabel_prf_jaccard_vehiclewise} reports the vehicle-wise multi-label fault detection performance of different methods in terms of Micro-Precision, Micro-F1, and Micro-Jaccard. 
Overall, VBFDD-Agent exhibits more stable and competitive performance across different vehicles. Averaged over all vehicles, VBFDD-Agent achieves Micro-Precision, Micro-F1, and Micro-Jaccard values of approximately 0.979, 0.954, and 0.919, respectively, outperforming the conventional multi-label learning baselines. 
This indicates that the proposed method can not only accurately identify activated alarm bits, but also maintain high consistency between predicted and ground-truth alarm combinations. From the vehicle-wise results, both VBFDD-Agent and conventional methods can achieve near-perfect or even completely correct detection results on vehicles with relatively simple fault patterns or regular label distributions. 
More importantly, when more complex concurrent fault combinations are present, such as in LB\_47 and LB\_74, the performance of conventional methods decreases noticeably, while the advantage of VBFDD-Agent becomes more evident.

\begin{figure*}[htbp]
  \centering
  \panel[0.45\linewidth]{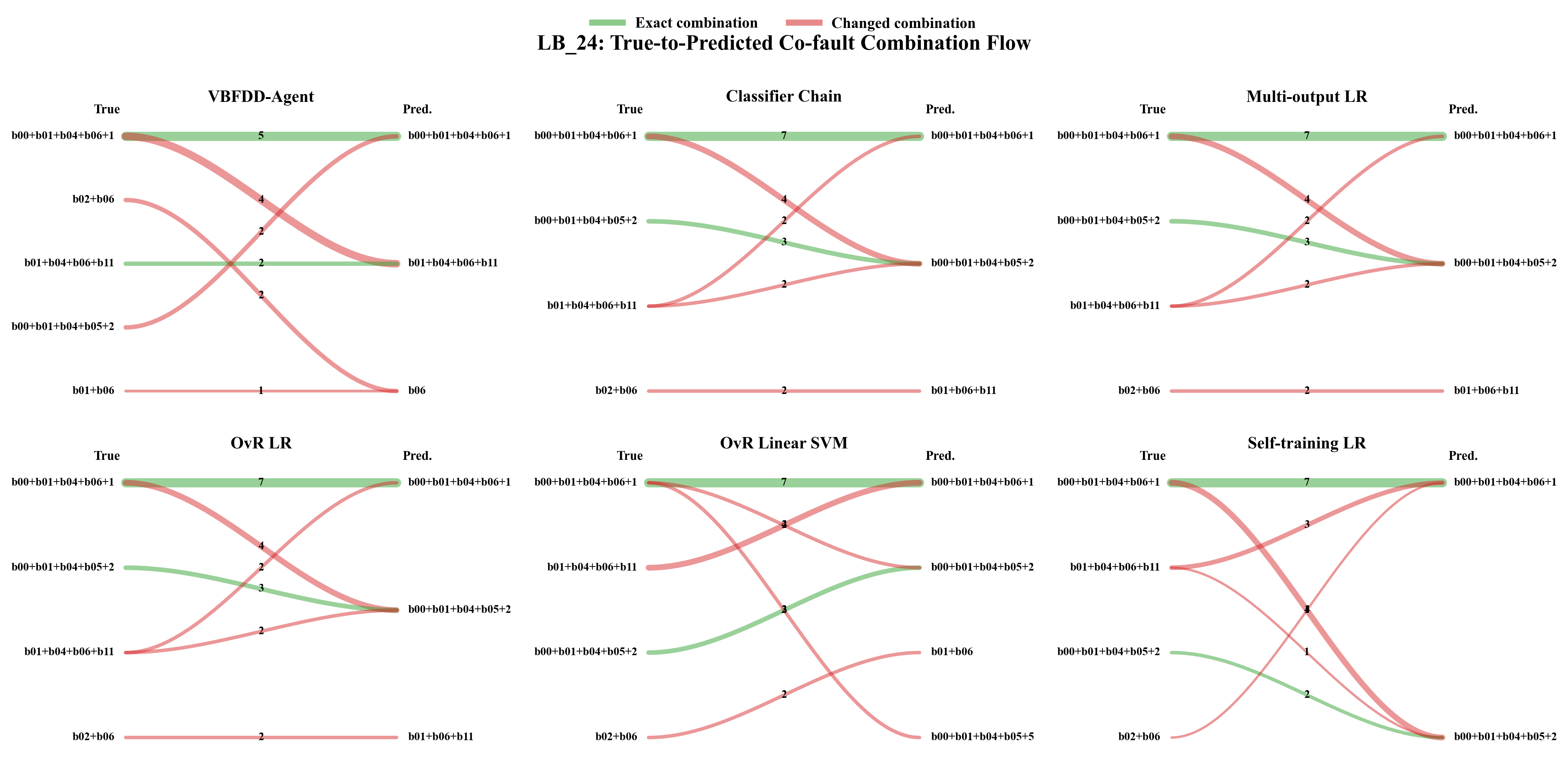}{(a)}\hfill
  \panel[0.45\linewidth]{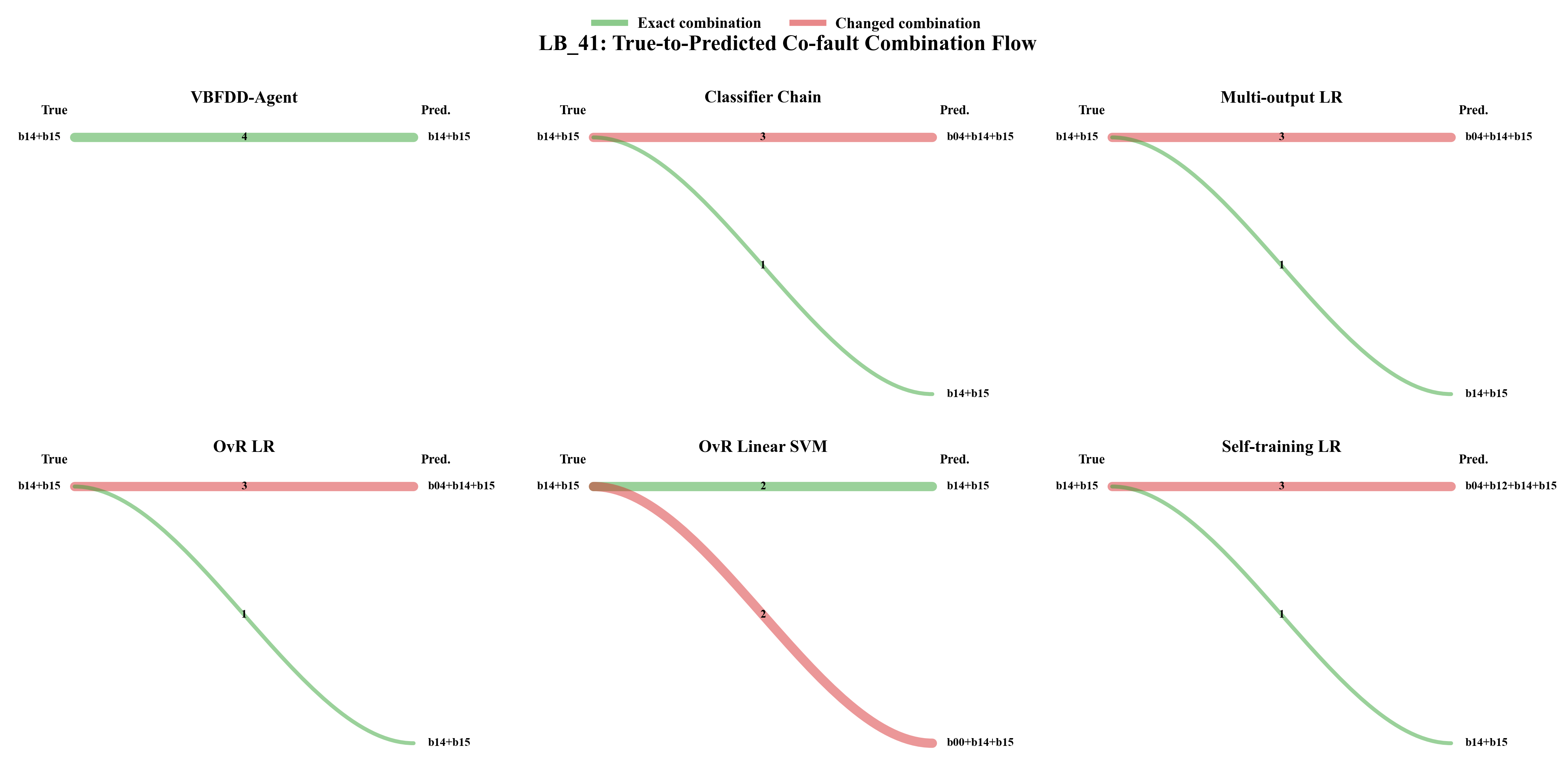}{(b)}\\
    \panel[0.45\linewidth]{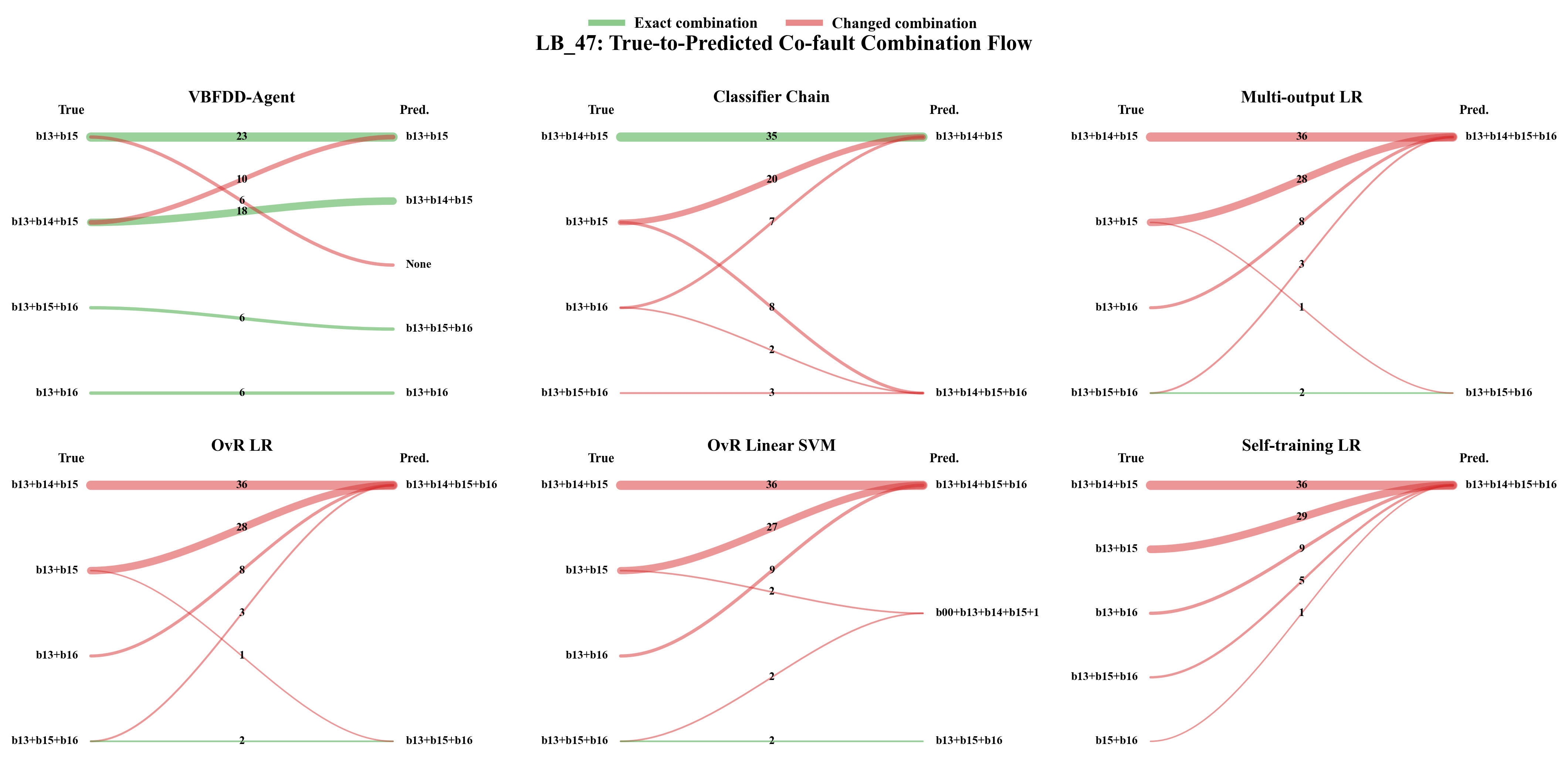}{(c)}\hfill
  \panel[0.45\linewidth]{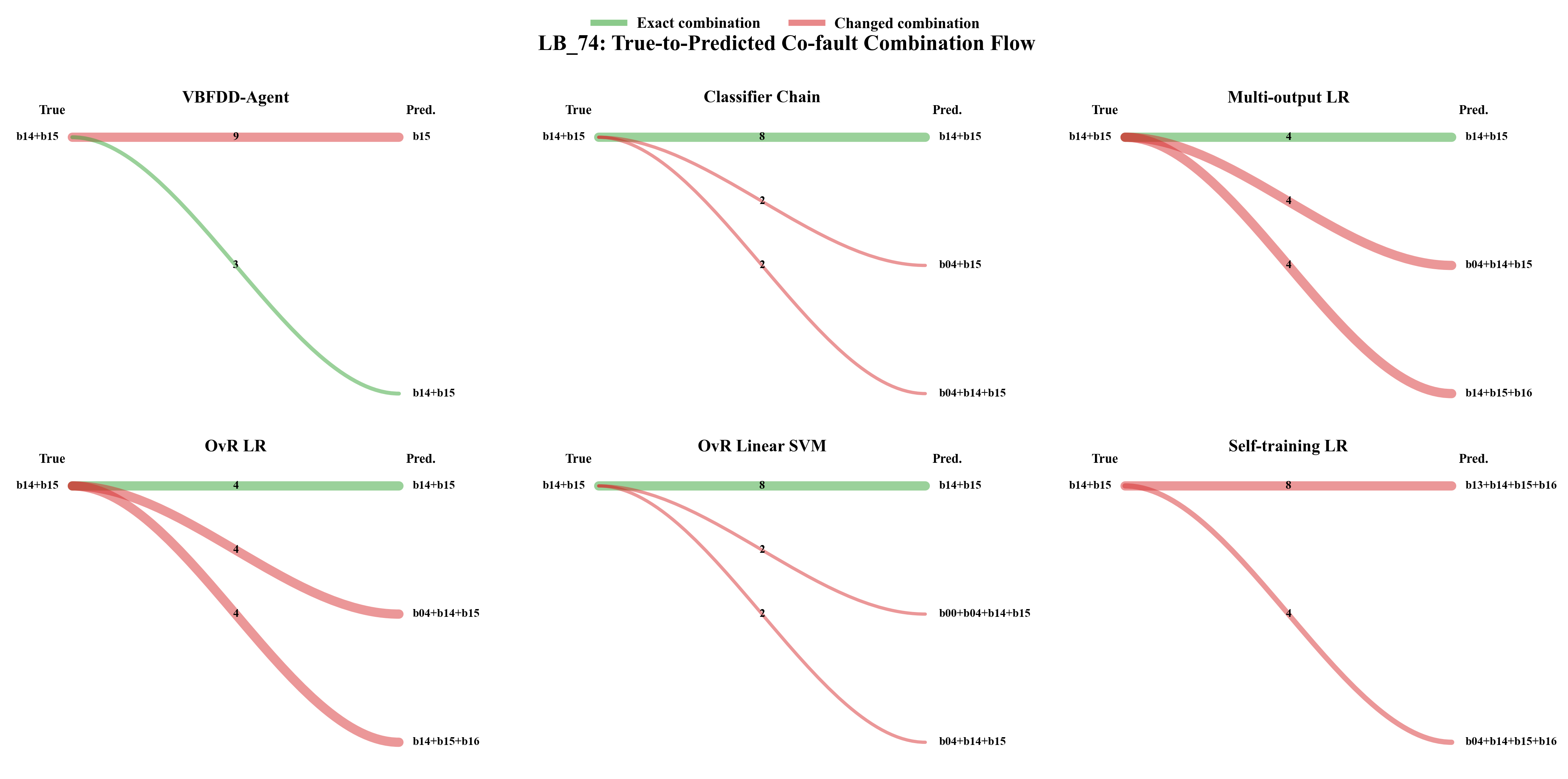}{(d)}\\
\caption{Flow diagrams of ground-truth and predicted co-occurring fault combinations for selected vehicles from LB\_24 to LB\_74. 
(a)--(d) show the combination-flow results of different vehicles, where the left side represents the ground-truth fault combinations and the right side represents the predicted alarm combinations. 
The width of each flow indicates the number of samples belonging to the corresponding true--predicted combination pair. 
A thick green flow from a ground-truth combination to the same predicted combination indicates that the model can accurately recover complete multi-fault combinations, whereas more red flows indicate that the model tends to predict concurrent faults as incomplete or mismatched combinations.}
\label{fig:combo_flow_grid}
\end{figure*}

Fig.~\ref{fig:combo_flow_grid} further illustrates the mapping relationships between ground-truth co-occurring fault combinations and model-predicted fault combinations. 
It can be observed that VBFDD-Agent produces more concentrated correct flows for samples with multiple co-occurring faults, where the major ground-truth fault combinations are mostly mapped to the same predicted combinations. 
This suggests that the proposed method can not only determine whether a sample is abnormal, but also effectively recover the combinational relationships among multiple fault labels. In contrast, conventional multi-label learning methods tend to produce more dispersed flows. 
Some ground-truth combinations are mapped to predicted combinations with missing or additional labels, indicating unstable combination-level recognition in co-occurring fault scenarios.

\begin{figure*}[htbp]
  \centering
  \panel[0.45\linewidth]{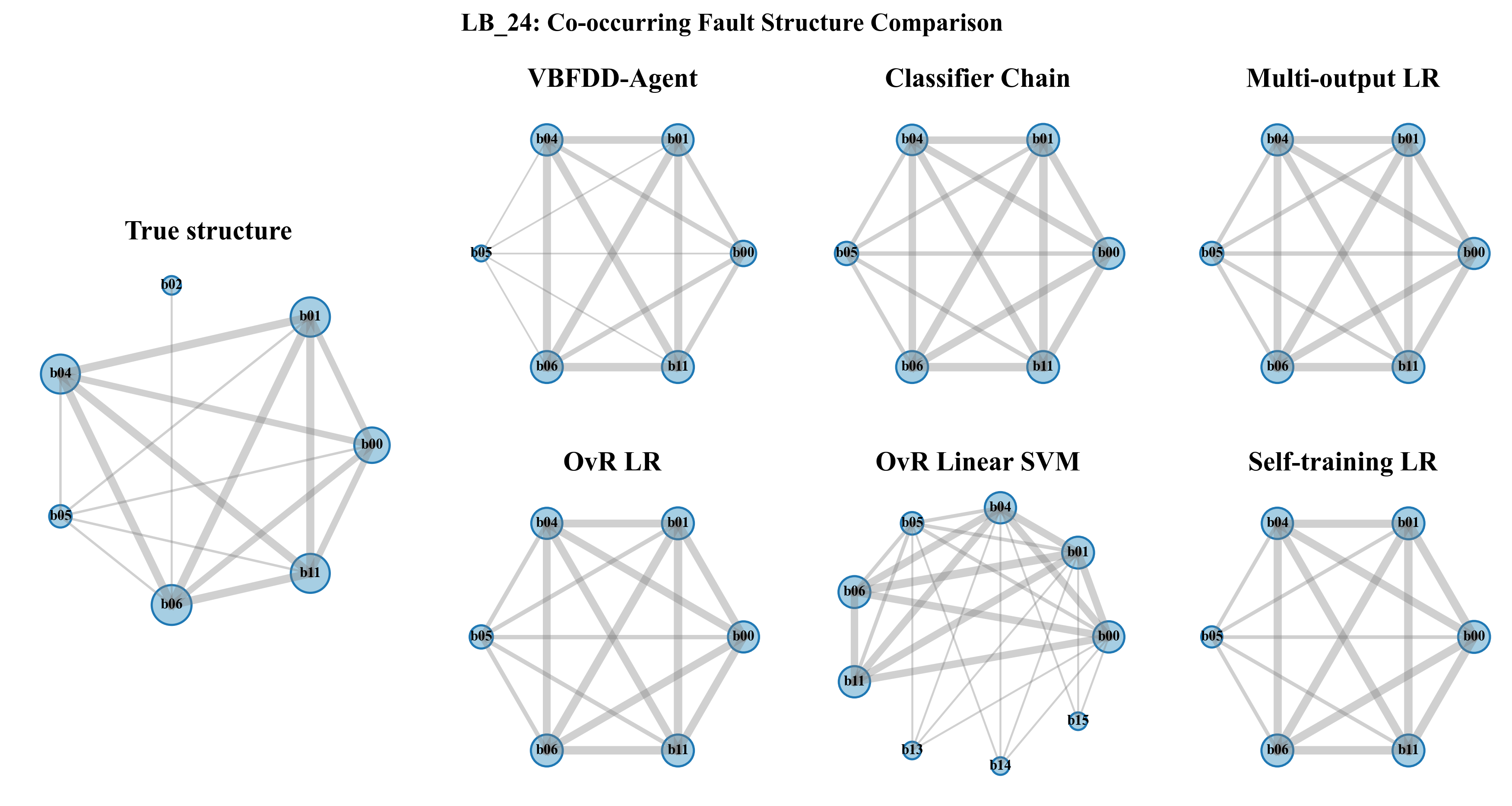}{(a)}\hfill
  \panel[0.45\linewidth]{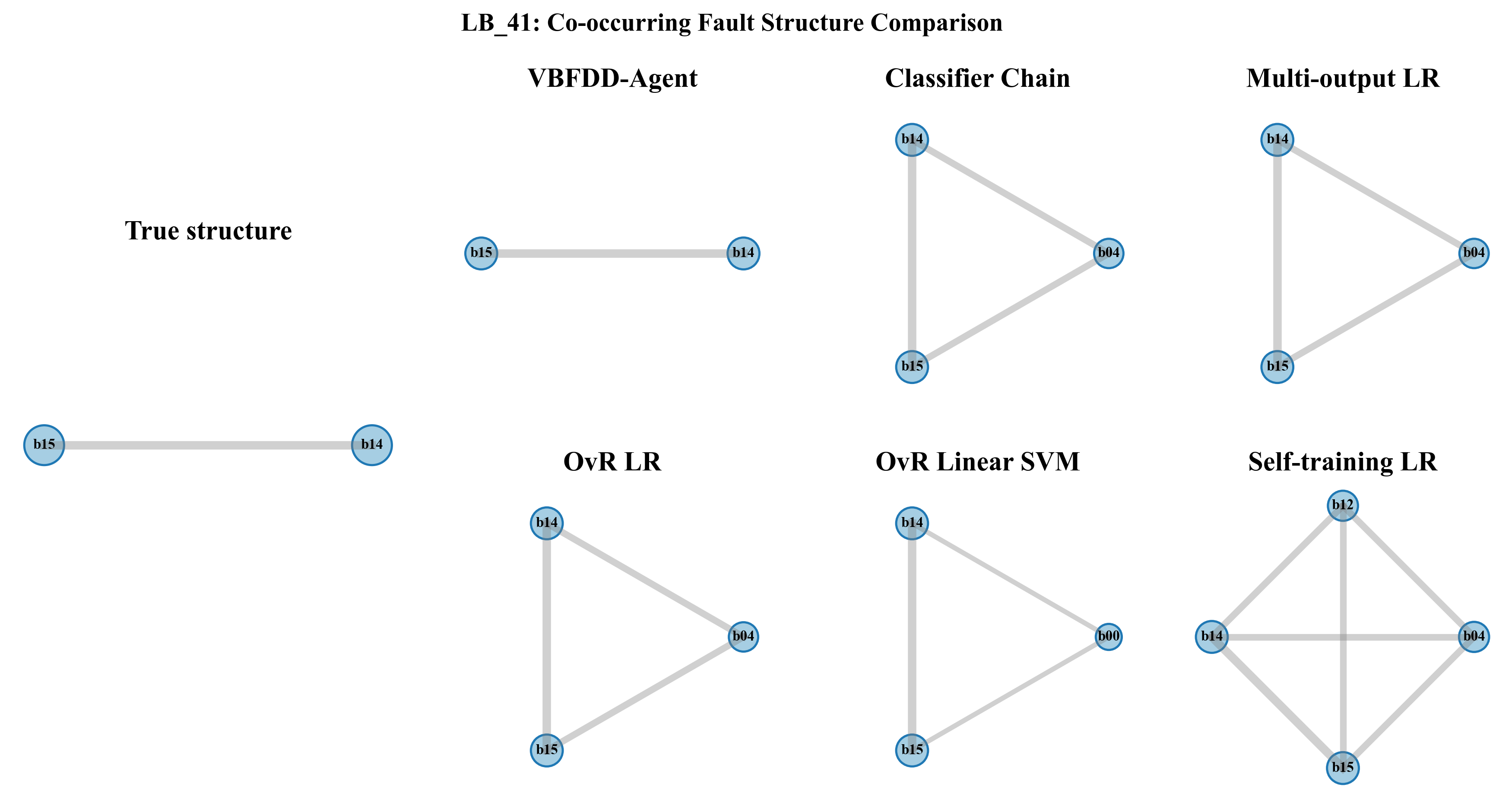}{(b)}\\
    \panel[0.45\linewidth]{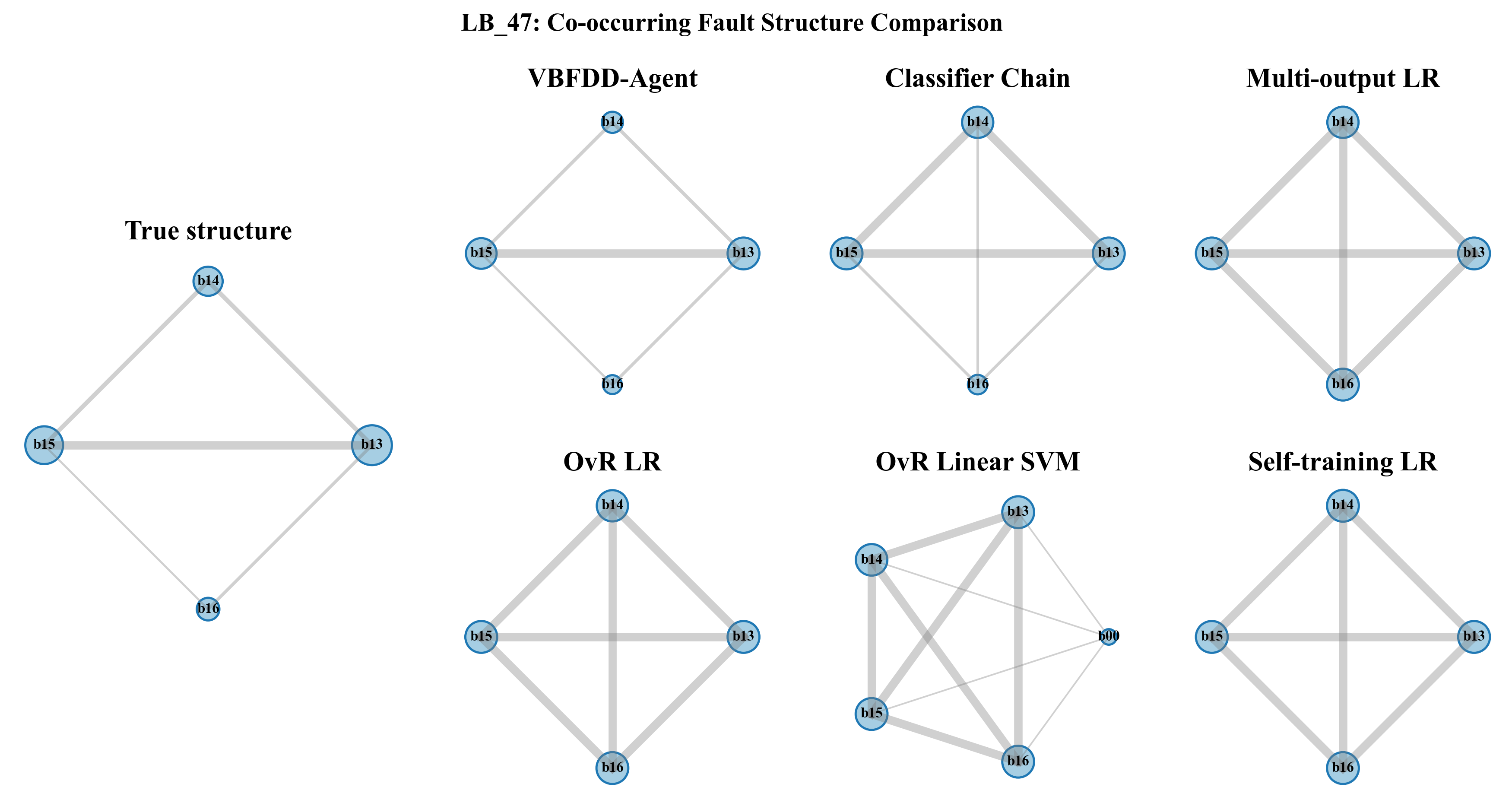}{(c)}\hfill
  \panel[0.45\linewidth]{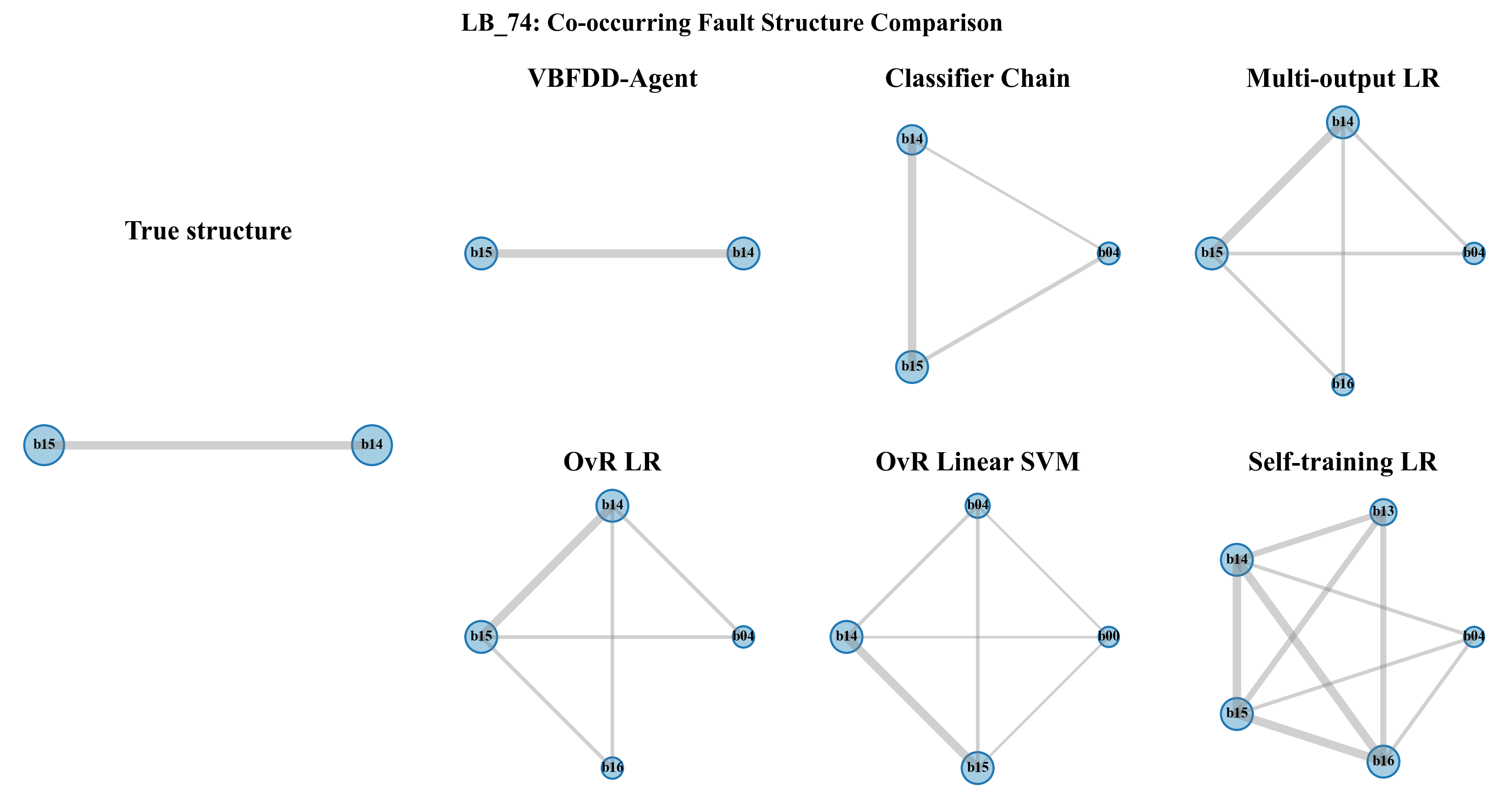}{(d)}\\
\caption{Co-occurrence structures of fault labels for selected vehicles from LB\_24 to LB\_74. 
(a)--(d) show the co-occurrence networks of different vehicles, where the left subgraph represents the ground-truth fault-label co-occurrence network and the right subgraph represents the predicted alarm-label co-occurrence network. 
Each node denotes a fault label, and each edge indicates that two fault labels occur simultaneously in the same sample. 
A larger node indicates that the corresponding label participates in co-occurrences more frequently, while a thicker edge indicates a higher co-occurrence frequency between two labels. 
A similar structure between the predicted and ground-truth networks suggests that the model can not only identify individual faults, but also capture the combinational relationships among concurrent faults.}
\label{fig:cooccurrence_structure}
\end{figure*}

Fig.~\ref{fig:cooccurrence_structure} compares the ground-truth fault-label co-occurrence structures with the predicted co-occurrence structures obtained by different methods. 
The ground-truth co-occurrence network characterizes the actual companion relationships among fault labels in the data, while the predicted co-occurrence network reflects whether the model can recover such label dependencies. 
From the overall network structure, the predicted co-occurrence network of VBFDD-Agent is closer to the ground-truth network, with major co-occurring nodes and high-frequency connections well preserved. 
This demonstrates that the proposed method can capture not only individual fault labels, but also the association patterns among fault labels. 
In comparison, some baseline methods exhibit missing edges, excessive weak connections, or structural shifts in their predicted networks.

Overall, the multi-label fault detection results verify the ability of VBFDD-Agent to understand concurrent battery faults. 
Compared with conventional numerical-feature-based multi-label classification methods, the proposed method represents battery operating states in a more semantically informative textual form through mechanism-informed descriptive text modeling, historical case retrieval, and evidence-grounded LLM reasoning. 
As a result, it can more effectively recover the combinational relationships among multiple concurrent alarm bits. 
Therefore, VBFDD-Agent not only improves multi-label fault detection performance, but also provides a more reliable diagnostic basis for subsequent fault interpretation, risk assessment, and maintenance recommendation generation.

    \section{Discussion}

\subsection{Expert Evaluation}

In addition to quantitative fault detection metrics, we further evaluate the practical usability of the maintenance recommendations generated by VBFDD-Agent. 
This evaluation is designed to go beyond the internal generation quality of the LLM and examine whether the generated disposal suggestions are reasonable from the perspective of real-world battery operation and maintenance. 
To this end, we invited three battery-domain experts who had no conflict of interest with this study to independently score the generated recommendations on a 100-point scale. 

\begin{table*}[!t]
\centering
\scriptsize
\caption{Expert evaluation of representative maintenance recommendations generated by VBFDD-Agent.}
\label{tab:expert_evaluation}
\renewcommand{\arraystretch}{1.25}
\resizebox{\textwidth}{!}{
\begin{tabular}{
>{\centering\arraybackslash}m{0.05\textwidth}
>{\raggedright\arraybackslash}m{0.23\textwidth}
>{\raggedright\arraybackslash}m{0.26\textwidth}
>{\raggedright\arraybackslash}m{0.36\textwidth}
>{\centering\arraybackslash}m{0.07\textwidth}
}
\toprule
\textbf{Case} 
& \textbf{Mechanism-informed state description} 
& \textbf{Disposal summary} 
& \textbf{Recommended actions} 
& \textbf{Avg. score} \\
\midrule

1 
& The vehicle is stationary with SOC close to 0\%. The total voltage is about 325.6 V, the load is light, and the insulation and temperature distribution are normal. However, the cell-voltage spread reaches about 84 mV, indicating pronounced voltage dispersion under low-SOC conditions.
& The battery is in a low-energy state with good insulation and uniform temperature, but voltage dispersion indicates possible cell degradation. Immediate charging is required to prevent deep-discharge damage. No critical safety alarms are present, but cell imbalance requires further diagnostic checks.
& Charge the battery immediately to raise SOC above 20\%. Perform a BMS diagnostic scan to check for weak or imbalanced cells. Monitor the cell-voltage spread during and after charging. Consider cell equalization or balancing if voltage dispersion persists. Inspect battery health and possible capacity fade due to high mileage.
& 95 \\

\midrule

2 
& The vehicle is stationary with SOC around 30\%. The total voltage is about 344.3 V, the load is light, cell-voltage consistency is good, insulation condition is good, and only a mild thermal gradient is observed. No prominent battery-oriented risk signature is identified from the signal description.
& The vehicle shows generally normal battery conditions. However, a drive motor controller temperature alarm is active, indicating a potential issue in the motor controller thermal management system rather than an immediate battery-related safety risk.
& Inspect the drive motor controller cooling system for blockages or malfunctions. Check temperature sensors and wiring related to the motor controller. Allow the vehicle to cool down and monitor whether the alarm persists after restart. Review historical records for recurring temperature alarms. If the alarm continues, perform diagnostic testing on the motor controller unit.
& 89 \\

\midrule

3 
& The vehicle is operating at a very low speed with SOC around 77\%. The total voltage is about 376.1 V, the estimated power is about 17.68 kW, and the current load is moderate. Cell-voltage dispersion and thermal gradient are both mild, while the insulation condition remains good.
& No alarms are triggered. The battery system is operating normally, and all monitored parameters remain within safe operating ranges.
& Continue routine monitoring of battery parameters. Maintain standard charging and usage practices. Schedule the next regular battery health inspection according to the maintenance plan.
& 90 \\

\bottomrule
\end{tabular}
}
\end{table*}

Table~\ref{tab:expert_evaluation} presents three representative cases, including a low-SOC voltage-dispersion case, a non-battery alarm case related to the drive motor controller, and a normal operating case. 
The average expert score of all three cases keeps high, indicating that the generated disposal summaries and recommended actions are highly recognized by domain experts. 
These results suggest that VBFDD-Agent can provide not only accurate fault detection results, but also practically meaningful maintenance guidance.

More importantly, the significance of this study is not limited to filling the gap in battery operation and maintenance corpora through signal-to-text modeling and LLM techniques. 
By embedding mechanism-informed information into textual representations, the proposed framework also provides a foundation for future text-signal collaborative modeling methods~\cite{jiang2026timexl}. 
This indicates a new paradigm for LLM-empowered battery health management, where numerical signals, mechanism knowledge, textual reasoning, and maintenance decision-making can be integrated into a unified framework.

\subsection{Trustworthy Diagnosis}

Although the experimental results demonstrate that VBFDD-Agent achieves competitive performance in binary anomaly detection and multi-label fault detection, the value of the proposed framework is not limited to high detection accuracy. 
In practical battery operation and maintenance scenarios, a diagnostic system is expected to provide not only fault labels or alarm scores, but also a trustworthy diagnostic chain that can be inspected, interpreted, and used by maintenance engineers. 
VBFDD-Agent addresses this requirement by connecting raw monitoring signals, mechanism-informed descriptive texts, historical case evidence, predicted alarm types, retrieved maintenance knowledge, and structured disposal recommendations into a unified reasoning process. 
This evidence-grounded design enables the diagnostic result to be traced back to both data-level evidence and knowledge-level evidence, thereby extending conventional FDD from isolated label prediction to interpretable decision support.

\begin{table}[!t]
\centering\scriptsize
\caption{Advantages and limitations of the training-free VBFDD-Agent framework.}
\label{tab:training_free_discussion}
\renewcommand{\arraystretch}{1.2}
\begin{tabular}{p{0.46\linewidth} p{0.46\linewidth}}
\toprule
\textbf{Advantages} & \textbf{Limitations} \\
\midrule

No task-specific retraining is required when applying the framework to new diagnostic cases. 
& Diagnostic performance depends on the coverage and quality of the historical case memory. \\

The framework can be updated by expanding the historical case memory and local maintenance knowledge base. 
& Maintenance recommendations depend on the completeness, reliability, and timeliness of the knowledge base. \\

It is suitable for fault-scarce and label-imbalanced scenarios, where supervised model training may be unstable. 
& Rare, unseen, or safety-critical fault patterns may still require expert verification. \\

The diagnostic output contains not only alarm predictions, but also evidence summaries and actionable maintenance recommendations. 
& The LLM reasoning process should be constrained by retrieved evidence to reduce unsupported or overly general responses. \\

It supports a traceable diagnostic chain from signal description to fault interpretation and disposal recommendation. 
& The quality of descriptive text modeling may influence retrieval accuracy and downstream reasoning reliability. \\

\bottomrule
\end{tabular}
\end{table}

Meanwhile, the proposed framework follows a training-free paradigm. Instead of retraining a task-specific model whenever new vehicle types, fault distributions, or maintenance rules emerge, VBFDD-Agent can be adapted by updating the historical case memory and local knowledge base. 
This makes it particularly suitable for fault-scarce, label-imbalanced, and continuously evolving operation and maintenance scenarios. 
Nevertheless, the training-free design also introduces several limitations, including dependence on the coverage of historical cases, the completeness of maintenance knowledge, and the need for expert verification under rare or safety-critical conditions. 
Therefore, VBFDD-Agent should be regarded as an evidence-grounded decision-support framework rather than a fully autonomous replacement for human experts. 
Its main contribution lies in moving battery FDD from high-accuracy prediction toward trustworthy, traceable, and actionable diagnosis.
    \section{Conclusion}

With the rapid development of electric vehicles, accurate, interpretable, and actionable battery fault detection and diagnosis has become increasingly important for safe and reliable operation. 
This paper proposes VBFDD-Agent, an LLM-empowered framework for electric vehicle battery fault detection, diagnosis, and maintenance recommendation generation. 
By transforming raw battery monitoring signals into mechanism-informed descriptive texts, the proposed method bridges numerical battery data and LLM-based semantic reasoning. 
On this basis, VBFDD-Agent integrates historical case retrieval, alarm-code prediction, local maintenance knowledge retrieval, and constrained LLM reasoning to produce structured diagnostic results and actionable disposal recommendations. 
Experiments on real-world data from ten electric vehicles show that the proposed method achieves competitive performance in vehicle-wise anomaly detection and demonstrates clear advantages in multi-label fault detection, especially for complex concurrent fault combinations. 
Expert evaluation further confirms the practical value and readability of the generated maintenance recommendations. 
Overall, this study provides a new paradigm for LLM-empowered battery FDD and offers a foundation for future text-signal collaborative modeling and intelligent battery operation and maintenance systems.
    \section*{Model and Data Availability}

The mechanism-informed descriptive text modeling results and the generated maintenance recommendation results are available at GitHub: \url{https://github.com/sjtu-chan-joey/VBFDD-Agent-Vehicle-Battery-Fault-Detection-and-Diagnosis-Agent}. 
The local knowledge base (LKB) was compiled from publicly available enterprise maintenance manuals, safety guidelines, technical documents, and vehicle specification sheets. 
The complete list of source URLs is provided in the GitHub repository and Supplementary Material.
    \section*{Acknowledgements}
This work is  Supported by Shanghai Jiao TongUniversity AI for Engineering Initiative.

\bibliographystyle{unsrt}
\bibliography{main.bib}

\end{document}